%% file: main.tex
\documentclass[10pt,twocolumn,letterpaper]{article}

\usepackage[pagenumbers]{iccv} % To force page numbers, e.g. for an arXiv version


\definecolor{iccvblue}{rgb}{0.21,0.49,0.74}
\usepackage[pagebackref,breaklinks,colorlinks,allcolors=iccvblue]{hyperref}
\usepackage{amsmath}
\def\paperID{14330} % *** Enter the Paper ID here
\def\confName{ICCV}
\def\confYear{2025}

\usepackage[accsupp]{axessibility}  
\usepackage{lipsum}
\usepackage{booktabs}
\usepackage{multirow}
\usepackage{amsmath}
\usepackage{array}
\usepackage{amssymb}
\usepackage{pifont} 
\usepackage[table]{xcolor}
\newcommand{\xmark}{\ding{55}}%
\newcommand{\failed}[1]
{\textcolor{Maroon}{#1}}
\newcommand{\success}[1]
{\textcolor{ForestGreen}{#1}}

\title{SuMa: A Subspace Mapping Approach for Robust and Effective Concept Erasure in Text-to-Image Diffusion Models}

\author{\textbf{Kien Nguyen}$^{1}$  \quad \quad \textbf{Anh Tran}$^{1}$ \quad \quad  \textbf{Cuong Pham}$^{2,1}$ \\
$^1$Qualcomm AI Research\\$^2$Posts and Telecommunications Institute of Technology (PTIT)\\
{\tt\small \{kienn, anhtra, pcuong\}@qti.qualcomm.com}
}

\begin{document}
\maketitle
\newcommand{\minisection}[1]{\vspace{2mm}\noindent{\textbf{#1}}}

\input{sec/0_abstract}    
\input{sec/1_intro}

\input{sec/2_related_work}

\input{sec/3_method}

\input{sec/4_Experiment}

{
    \small
    % \bibliographystyle{ieeenat_fullname}
    % \bibliography{main}

}

\appendix

\input{sec/X_sup.tex}

\end{document}

%% file: sec/0_abstract.tex
\begin{abstract}
The rapid growth of text-to-image diffusion models has raised concerns about their potential misuse in generating harmful or unauthorized contents. To address these issues, several Concept Erasure methods have been proposed. However, most of them fail to achieve both robustness, i.e., the ability to robustly remove the target concept., and effectiveness, i.e., maintaining image quality. While few recent techniques successfully achieve these goals for NSFW concepts, none could handle narrow concepts such as copyrighted characters or celebrities. Erasing these narrow concepts is critical in addressing copyright and legal concerns. However, erasing them is challenging due to their close distances to non-target neighboring concepts, requiring finer-grained manipulation. In this paper, we introduce Subspace Mapping (SuMa), a novel method specifically designed to achieve both robustness and effectiveness in easing these narrow concepts. SuMa first derives a target subspace representing the concept to be erased and then neutralizes it by mapping it to a reference subspace that minimizes the distance between the two. This mapping ensures the target concept is robustly erased while preserving image quality. We conduct extensive experiments with SuMa across four tasks: subclass erasure, celebrity erasure, artistic style erasure, and instance erasure and compare the results with current state-of-the-art methods. Our method achieves image quality comparable to approaches focused on effectiveness, while also yielding results that are on par with methods targeting completeness.

\end{abstract}

%% file: sec/1_intro.tex
\section{Introduction}
\label{sec:intro}

Text-to-image models (T2I) at a large scale have achieved notable success in producing lifelike images \cite{ramesh2022hierarchical, nichol2022glide,rombach2022high,saharia2022photorealistic,yuscaling,chang2023muse}. This progress is not only a result of enhanced algorithms and computing resources but also the extensive datasets gathered from the Internet \cite{schuhmann2022laion}. Unfortunately, these datasets often contain copyrighted works, artistic creations \cite{jiang2023ai,somepalli2023diffusion,shan2023glaze}, NSFW content, and personal images, raising ethical and legal concerns \cite{lu2024mace}. 

Some publicly available models, like Stable Diffusion (SD) \cite{RobinSD1.4}, trained on vast web-scraped datasets, can generate inappropriate content, including copyrighted images, offensive material, and deepfakes, raising serious community concerns. To address this, researchers have attempted to filter out such content, resulting in the release of Stable Diffusion 2.0 \cite{RobinSD2.0}. However, this version, despite the filtering efforts, delivers lower image quality than its predecessor, with some explicit content still slipping through. Another approach involves post-processing filters, such as NSFW detectors, which flag inappropriate outputs and replace them with blank images. Yet, these safeguards can be easily bypassed once the model is publicly accessible \cite{lu2024mace}.

Recent research has concentrated on fine-tuning text-to-image (T2I) diffusion models to ‘forget’ inappropriate content, giving rise to Concept Erased Models (CEMs) as a practical approach to mitigate such issues. The first generation of these models, which we term Effective Concept Erased Models (E-CEMs), demonstrates success in removing inappropriate content while maintaining image quality comparable to the original outputs \cite{gandikota2023erasing, kumari2023ablating, lyu2024one, lu2024mace, bui2024removing}. However, recent studies have uncovered significant vulnerabilities in these methods \cite{pham2023circumventing, zhang2025generate, tsai2023ring}. For instance, UnlearnDiff \cite{zhang2025generate} identifies combinations of existing tokens within current models capable of reconstructing the target concept, effectively bypassing the erasing mechanism. Additionally, CCE \cite{pham2023circumventing} leverages Textual Inversion \cite{gal2022image} to derive a new token that regenerates the erased concept. Collectively, these findings suggest that E-CEMs fail to robustly eliminate target concepts from the model’s latent space, leaving them susceptible to regeneration through adversarial techniques.

The second generation, which we call Robust Concept Erased Models (R-CEMs), addresses vulnerabilities revealed by adversarial attacks. Early approaches, like RACE \cite{kim2024race}, use an iterative framework based on E-CEM, identifying adversarial tokens and erasing them. AdvUnlearn \cite{zhang2024defensive} improves RACE by adding a prior preservation loss to retain model utility. However, while these methods mitigate UnlearnDiff attacks, they struggle against CCE. This is because CCE generates new tokens, while UnlearnDiff relies on existing token combinations, limiting its versatility compared to CCE's broader search space. We provide a discussion about these two attacking methods in Appendix \ref{sec:Related_Work_Discussion}. More recent methods, such as STEREO \cite{srivatsan2025stereo}, refine this process by identifying and erasing all adversarial tokens detected by CCE simultaneously. While it demonstrates notable success in terms of robustness and resilience against current attack methods, it incurs a significant reduction in image quality. In contrast, DUO \cite{parkdirect} first applies an image editing approach to modify inappropriate content, then leverages the Diffusion-DPO framework \cite{wallace2024diffusion} to assign high scores to appropriate images and low scores to inappropriate ones while also introducing a novel prior preservation loss. This method successfully protects the model from existing attacks while maintaining image quality. However, it is currently limited to erasing NSFW content.

While NSFW content is a common target of concept erasure, narrow concepts are not less important targets. Here, we use the terms ``narrow concepts'' to refer to fine-grained categories or specific instances within, such as ``English Springer'', ``Grumpy Cat'', or ``Elon Musk''. The ability to erase narrow concepts is critical, as they may involve copyright issues or contribute to the spread of misleading information. While DUO is the only robust and effective concept erasure technique, it struggles to handle narrow concepts.
% such as copyrighted characters (e.g., Mickey Mouse), artistic styles, or celebrities. 
We discuss in more detail the reason in the Appendix \ref{sec:Related_Work_Discussion}. In short, this method relies on an editing approach to erase concepts. For instance, it edits the ``nudity" concept to ``dressed in", where the difference between the two is clear. However, for narrow concepts, such as editing ``Grumpy Cat" to ``Cat", it is harder to classify these as distinct classes, requiring a finer-grained method to address. 
% However, the ability to erase narrow concepts is critical, as they may involve copyright issues or contribute to the spread of misleading information.

Given the aforementioned challenges, our paper presents complementary work to DUO \cite{parkdirect}, focusing on achieving a balance between robust and effective erasure for narrow concepts. We first investigate why adversarial tokens can bypass E-CEMs and analyze how STEREO achieves robustness against such attacks. We then introduce our method, Subspace Mapping Concept Erasure (SuMa). Our approach constructs both a target subspace containing the concept to be erased and a reference subspace that is independent yet close to the target subspace by identifying their basis vectors. We eliminate the target subspace by fine-tuning the model to optimize a novel objective function, aiming to project the basis of the target subspace onto the reference subspace. This method is integrated with E-CEM techniques, making the model robust against adversarial attacks while still preserving image quality. As a result, SuMa achieves image quality on par with E-CEMs in terms of \textit{FID} and \textit{CLIP-score}, and an \textit{Attack Success Rate (ASR)} comparable to the state-of-the-art R-CEM method, STEREO. Our main contributions are as follows:
\begin{itemize} 
\item We analyze why adversarial tokens can bypass E-CEMs and how STEREO remains resilient against these attacks. 
\item We propose the Subspace Mapping Concept Erasure (SuMa) framework, which balances robustness and effectiveness in erasing narrow concepts. Our framework first constructs a target subspace that captures the concept to be erased and a reference subspace that is distinct yet proximate to the target. We then introduce a novel objective function that removes the target concept by mapping the target subspace onto the reference subspace.
\item Extensive experiments across four tasks—object erasure, celebrity erasure, artistic style erasure, and subclass erasure—demonstrate that SuMa offers the best trade-off between completeness and effectiveness. SuMa completely removes the target concept while preserving model utility. To the best of our knowledge, we are the first to achieve these dual objectives for narrow concepts. \end{itemize}

\begin{figure}[t]
  \centering
   \includegraphics[width=1.0\linewidth]{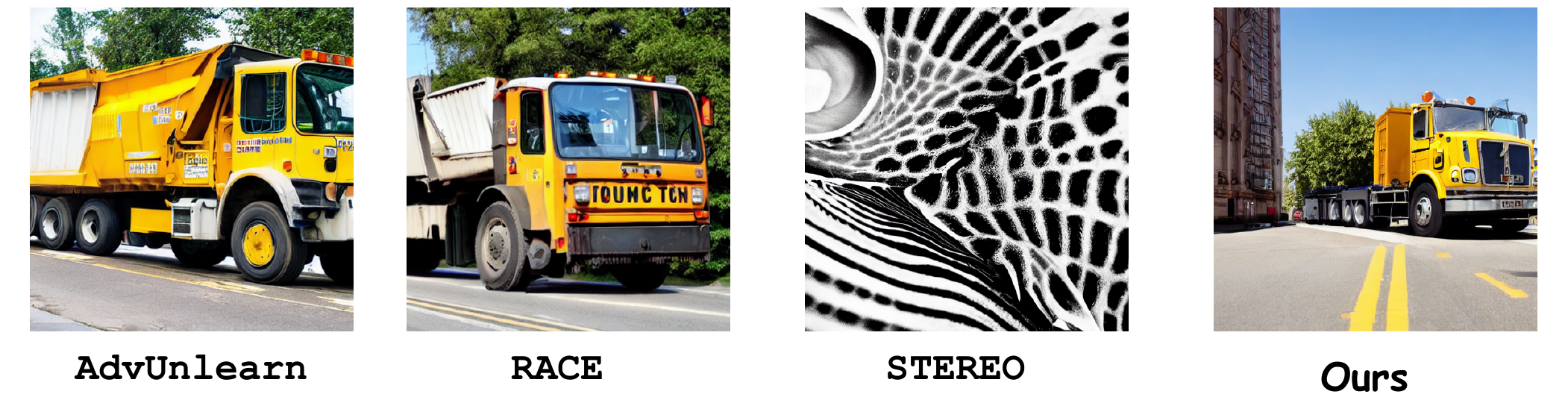}
   \vspace{-5mm}
   \caption{Illustration of current C-CEMs methods trained to erase \textit{garbage truck} against the CCE Attack. We can see that AdvUnlearn and RACE are still vulnerable to the CCE Attack, while STEREO successfully defends against it, though at the cost of image quality. Our method successfully balances Completeness and Effectiveness, changing \textit{garbage truck} into \textit{Semi-trailer truck.}} 
   \label{fig:Illustrate}
   \vspace{-1em}
\end{figure}

%% file: sec/2_related_work.tex
\section{Related Work}

\begin{figure*}[t]
  \centering
  \begin{subfigure}{0.32\linewidth}
    \includegraphics[width=1.0\linewidth]{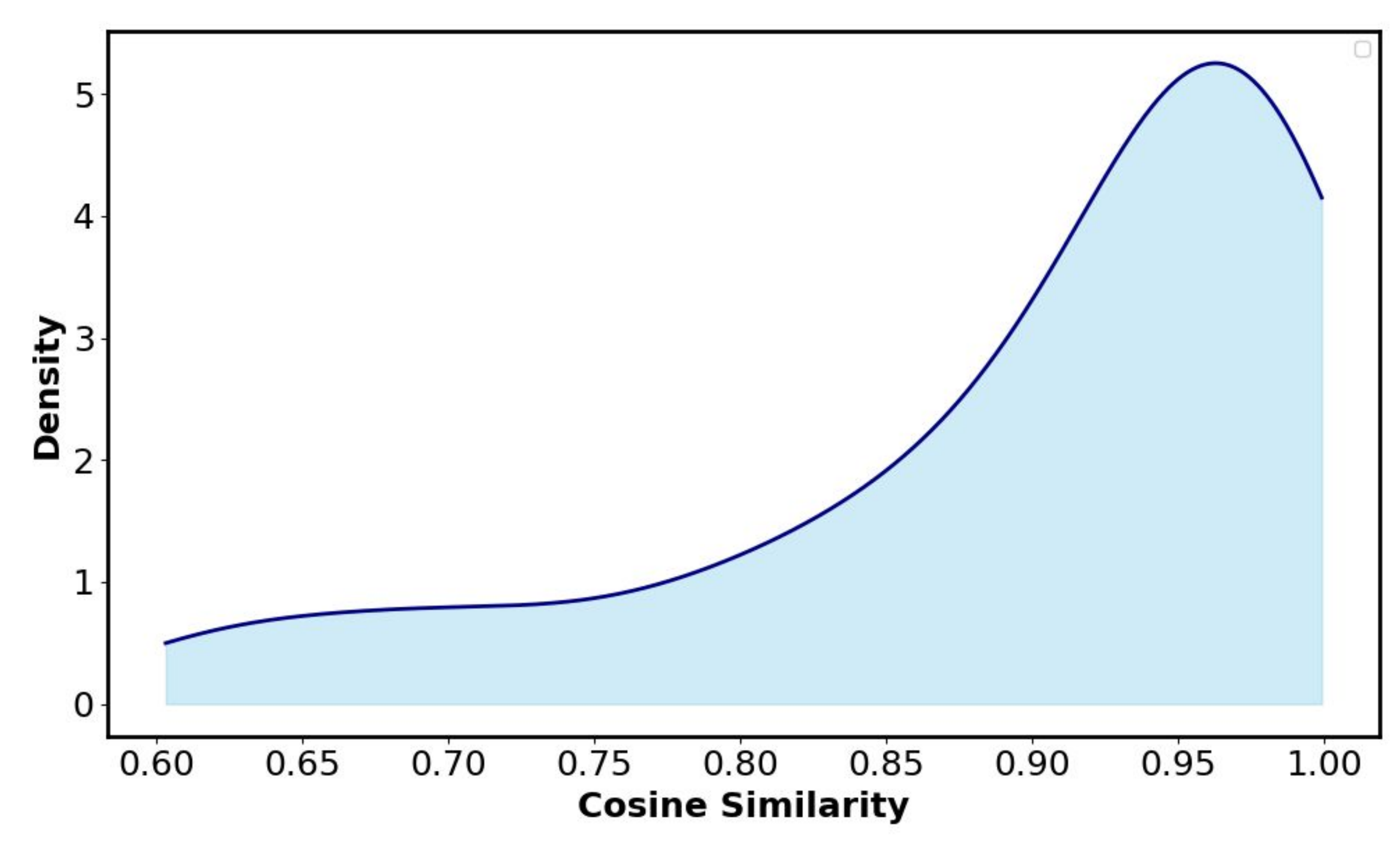}
    \caption{Cosine similarity of TIs from SD1.4.}
    \label{fig:cosin-same-model}
  \end{subfigure}
  \hfill
  \begin{subfigure}{0.32\linewidth}
    \includegraphics[width=1.0\linewidth]{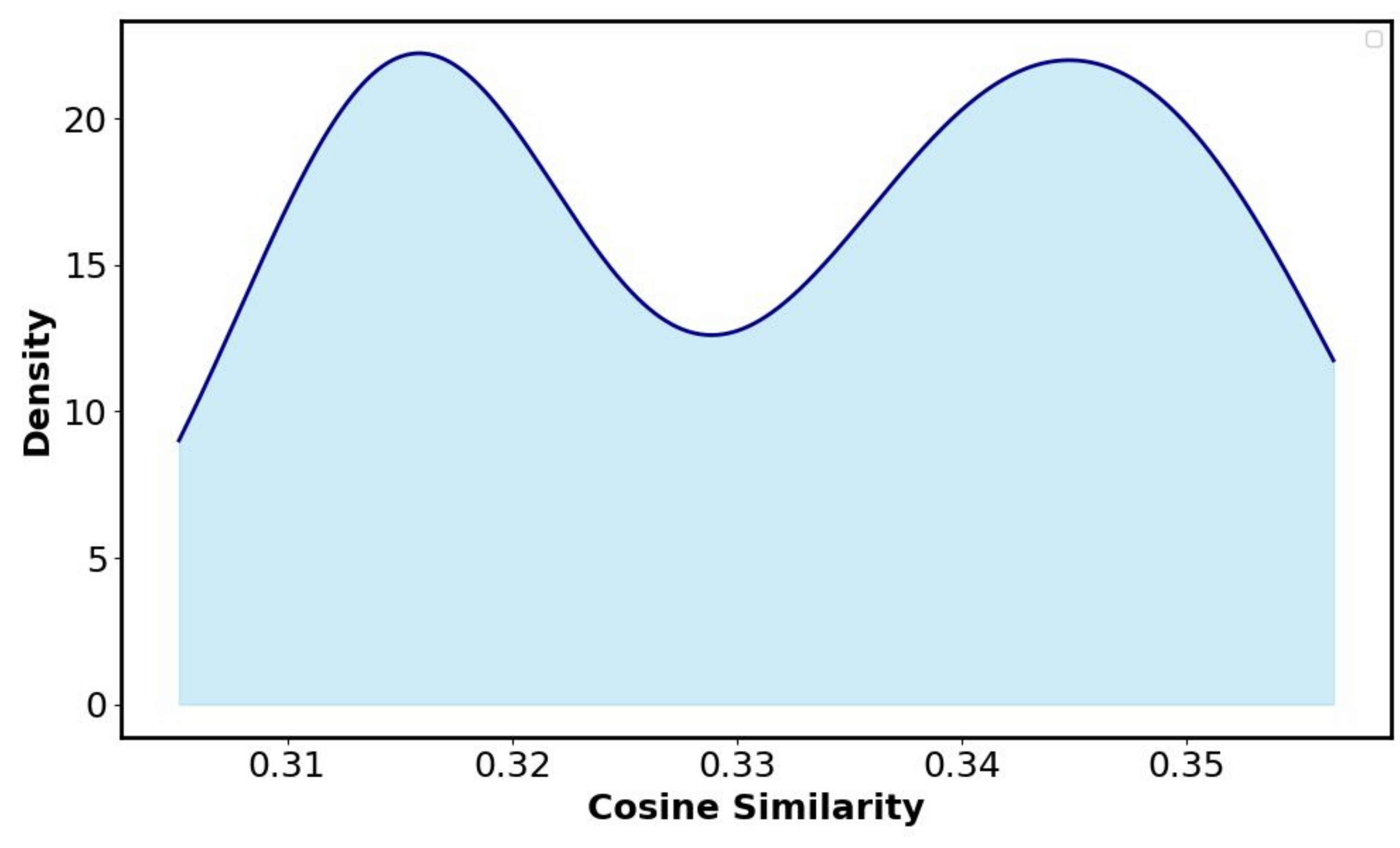}
    \caption{Cosine similarity of TIs at different CA levels.}
    \label{fig:cosin-diff-model}
  \end{subfigure}
  \hfill
  \begin{subfigure}{0.32\linewidth}
    \includegraphics[width=0.9\linewidth]{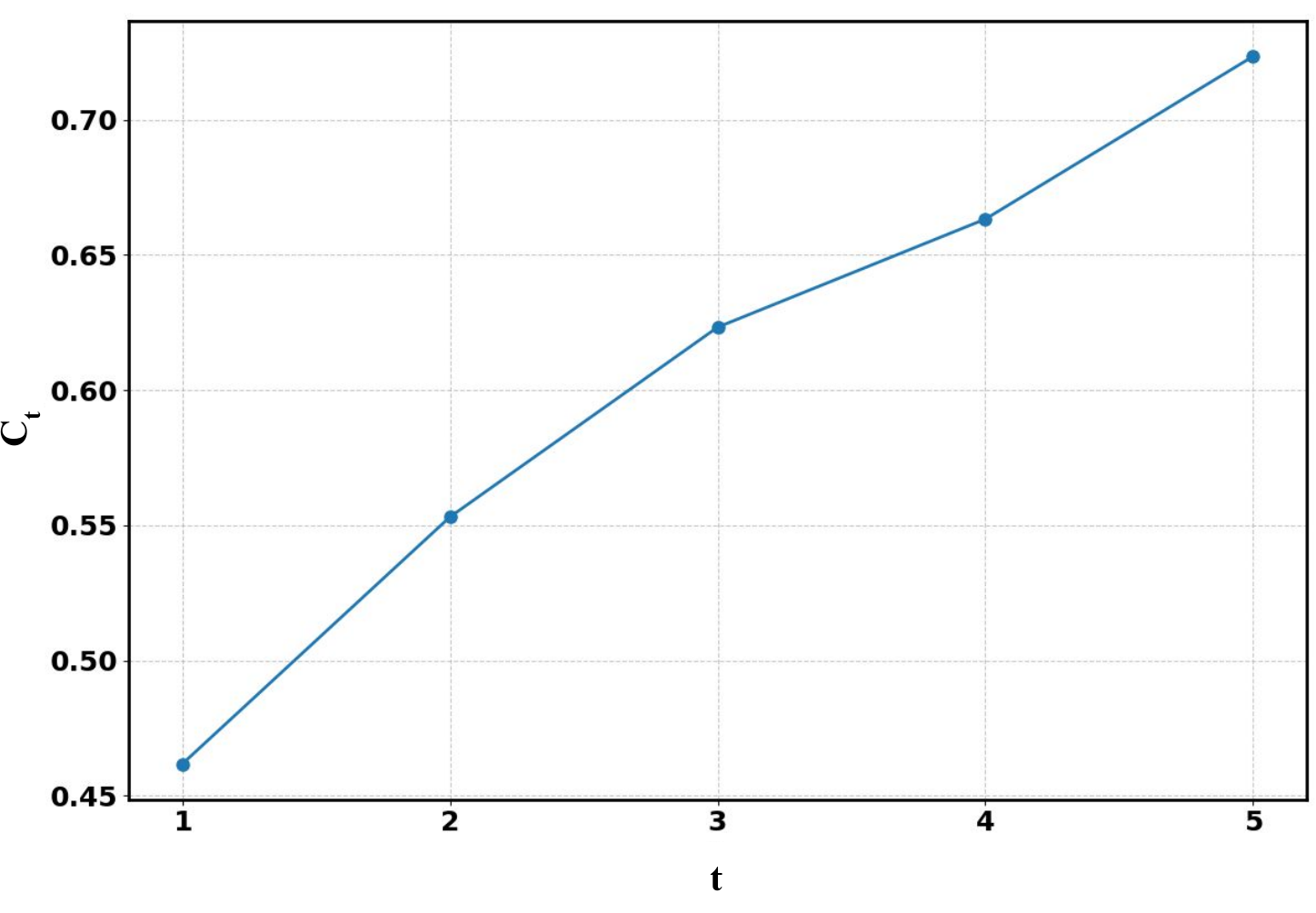}
    \caption{Cosine similarity of TIs with subspace.}
    \label{fig:cosin-subspace}
  \end{subfigure}
  \vspace{-2mm}
  \caption{Distribution of cosine similarity scores among TIs in three cases: from the same SD1.4 but different TI training steps, different erasure levels of CA, and between a TI at erasure level $t+1$ and the subspace created by the $t$ previous TIs.}
  \vspace{-2mm}
  \label{fig:short}
\end{figure*}
% \subsection{Text-to-Image Diffusion Model}
% Latent Diffusion Models \cite{rombach2022high} learn to reverse a Markov chain process where noise is gradually added to the latent representation of the input image at multiple timesteps \( t \in [0, T] \). Instead of working directly on the image space, these models operate within a lower-dimensional latent space, which is more computationally efficient. The noisy latent variable \( z_t \) at time-step \( t \) can be computed as \( z_t = \sqrt{\alpha_t} z_0 + \sqrt{1 - \alpha_t} \epsilon \), where \( z_0 \) represents the clean latent, \( \epsilon \) is random noise, and \( \alpha_t \) is the coefficient that determines the noise strength. The denoising network \( \Phi(z_t, \mathcal{T}(c), t) \) is trained to predict the noise added to \( z_{0} \) to obtain \( z_t \), conditioned on the encoded prompt \( \mathcal{T}(c) \), where \( \mathcal{T} \) is a text-encoder network and \( c \) is the text prompt. This conditioning enables control over the generated output based on the given prompt. The training objective is as follows:
% \begin{equation}
%     \mathcal{L} = \mathbb{E}_{z_t, \epsilon, c, t} \left[ \| \epsilon - \Phi(z_t, \mathcal{T}(c), t) \|_2^2 \right]
%     \label{eq: Diff_Equal}
% \end{equation}
\subsection{Effective Concept Erasure (E-CEMs)}
The first generation of the Concept Erasure Models (CEMs) was designed to fine-tune original models to forget inappropriate content, successfully removing unwanted elements while closely maintaining image quality. Concept Ablation (CA) \cite{kumari2023ablating} finetunes the model to redirect the target prompt, e.g., \textit{``grumpy cat"}, to be similar as a general term, e.g., \textit{``cat"}, by aligning the noise predictions on these prompts. ESD \cite{gandikota2023erasing} pushed the noise predictions on the target concept away from the original noise predictions by reversing the direction of conditional guidance \cite{ho2022classifier}. SPM \cite{lyu2024one} focused on preserving non-target concepts during erasure by applying the CA method with added regularization that maintained noise predictions on the closet concept in the CLIP Embedding Space. MACE \cite{lu2024mace} proposes a LORA-Finetuning \cite{hu2021lora} method to erase multiple concepts simultaneously. Despite these advancements, recent adversarial attacks \cite{pham2023circumventing,zhang2025generate,tsai2023ring} successfully recovered the ``erased" concepts, exposing vulnerabilities in all these methods and indicating a need for further improvements in concept erasure techniques.
\subsection{Circumventing Concept Erasure} \label{sec:Circumventing}
Several methods have been proposed to attack E-CEMs \cite{pham2023circumventing, zhang2025generate, tsai2023ring}. P4D \cite{chin2023prompting4debugging} enhances adversarial prompts by ensuring that noise estimates between pre-trained and concept-erased models are similar. UnlearnDiff \cite{zhang2025generate} is similar to P4D but leverages the classification capabilities of diffusion models, removing the reliance on pre-trained models. Ring-A-Bell \cite{tsai2023ring} creates adversarial prompts by guiding a “safe” prompt toward an “unsafe” one in the text vector space. Among these, CCE \cite{pham2023circumventing} is the simplest and most effective, employing Textual Inversion (TI) \cite{gal2022image} to derive a new token from the E-CEM, which regenerates the erased concept. Unlike UnlearnDiff, which relies on combinations of existing tokens, CCE searches for a new token in the latent text space, greatly widening its search space. Hence, defending against CCE is far more challenging than against UnlearnDiff.

\subsection{Robust Concept Erasure (R-CEMs)}
Motivated by the need to protect CEMs from adversarial attacks, several methods have emerged as part of the second generation of CEMs \cite{kim2024race, srivatsan2025stereo, zhang2024defensive}. RACE \cite{kim2024race} proposes an iterative framework that first identifies adversarial tokens using the UnlearnDiff framework but with faster optimization. It then erases the found tokens using a loss function from ESD, providing protection against UnlearnDiff. However, as shown in Fig. \ref{fig:Illustrate}, it remains vulnerable to more advanced CCE attacks. AdvUnlearn \cite{zhang2024defensive} is similar to RACE, but it introduces a prior preservation loss during the erasing phase to maintain the model’s utility. Still, it is susceptible to CCE attacks, as illustrated in Fig. \ref{fig:Illustrate}. STEREO \cite{srivatsan2025stereo} successfully defends against all existing attack methods by iteratively identifying all Textual Inversions (TIs) from the concept-erased model, and erasing all of them simultaneously from the original model. However, this method leads to a significant degradation in image quality, resulting in an FID nearly double that of E-CEMs. DUO \cite{parkdirect} addresses both robustness and effectiveness by leveraging Diffusion-DPO to assign high scores to concept-edited images and low scores to inexact ones. It also introduces a novel prior preservation loss to preserve the model’s utility. However, this approach is currently limited to NSFW concepts.

%% file: sec/3_method.tex
\section{Method}
\begin{figure*}[t]
  \centering
  \begin{subfigure}[t]{0.49\linewidth} % Reduced width slightly for better spacing
    \includegraphics[width=.98\linewidth]{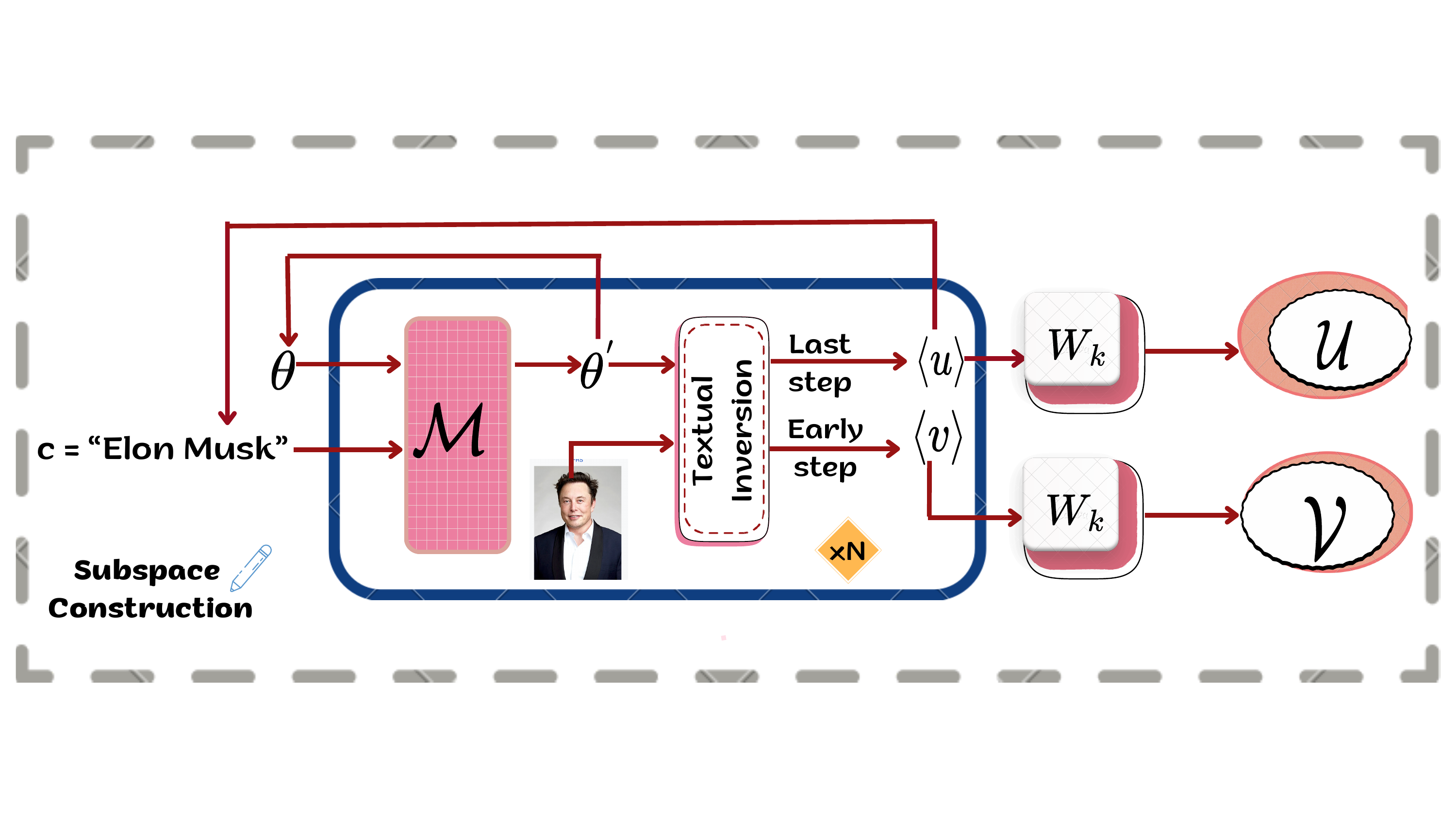} % Include Stage1 image
    \vspace{-2em}
    \caption{\textbf{Subspace Construction}: Following the STEREO approach, we iteratively remove the Target Concept from model $\theta$ using method $\mathcal{M}$, then run Textual Inversion (TI) and extract the corresponding tokens. At each iteration, we take the token from the final TI step and one from an early step to form the basis of the \textit{Target Subspace} and \textit{Reference Subspace}, respectively.}
    \label{fig: Stage1}
  \end{subfigure}%
  \hspace{0.01\linewidth} % Adjusted spacing between subfigures
  \begin{subfigure}[t]{0.49\linewidth} % Reduced width slightly for better spacing
    \includegraphics[width=.98\linewidth]{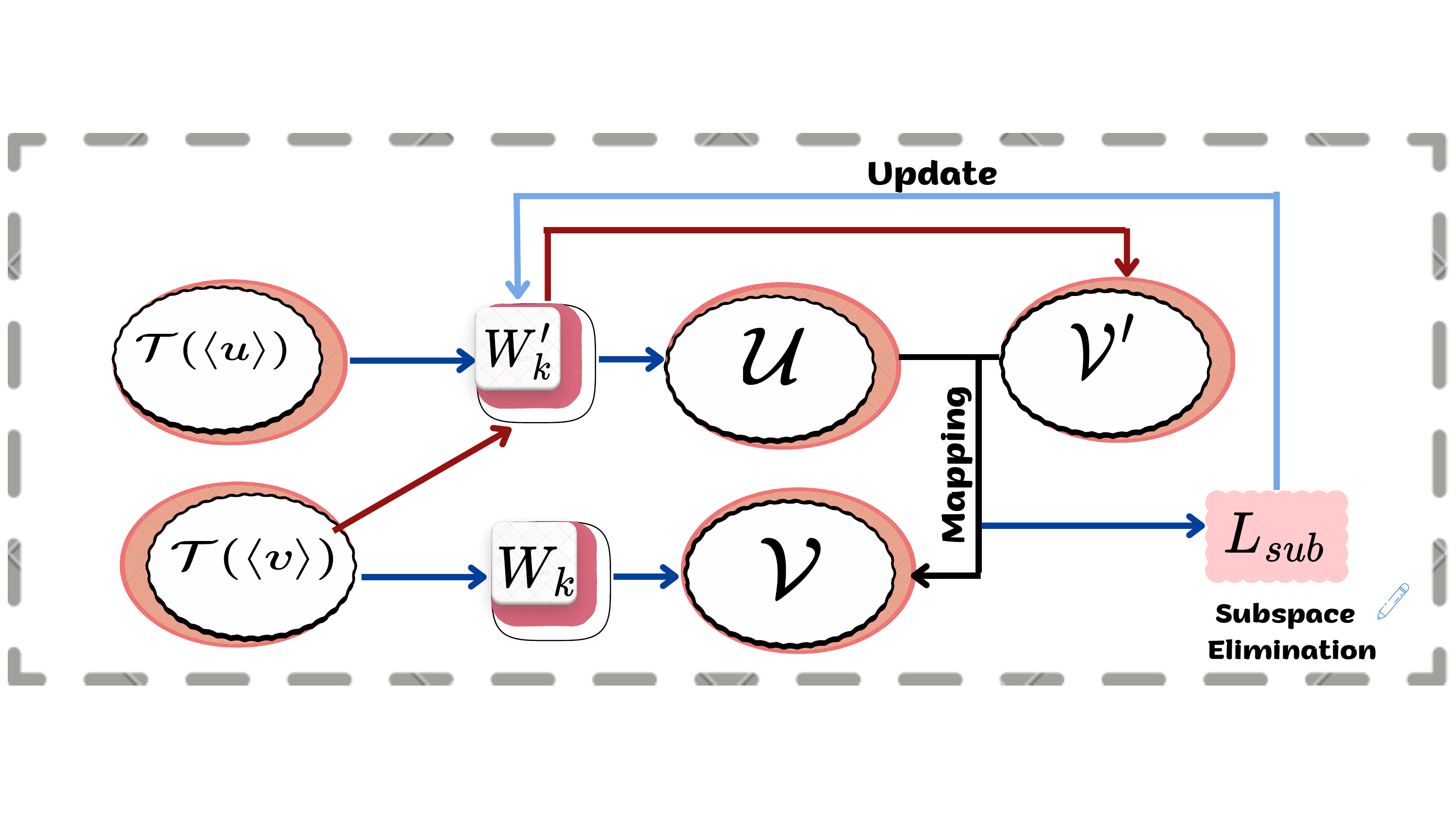} % Include Stage2 image
    \vspace{-2em}
    \caption{\textbf{Subspace Elimination}: In this stage, our objective is to map the \textit{Target Subspace} $\mathcal{U}$ onto the \textit{Reference Subspace} $\mathcal{V}$, and to maintain the \textit{Reference Subspace} in its fine-tuned state $\mathcal{V'}$ unchanged. We let $\mathcal{T}$ represent the Text-encoder and $L_{sub}$ denote the cost of mapping $\mathcal{U}$ to $\mathcal{V}$ and $\mathcal{V'}$ to $\mathcal{V}$. The parameter $W_k^{\prime}$
    is initialized as $W_k$ and will be updated during fine-tuning. }
    \label{fig: Stage2}
  \end{subfigure}
  \caption{Our proposed framework consists of two main stages. The first stage, \textbf{Subspace Construction}, is used to build Target Subspace and Reference Subspace. The second stage, \textbf{Subspace Elimination}, aims to eliminate the Target Subspace.}
  \label{fig:Main_Framework}
  \vspace{-1em}
\end{figure*}

In this section, we begin by exploring whether E-CEMs, trained to erase a single token, can also erase tokens that differ in form but convey the same concept (\cref{sec:analysis}). This analysis sets the stage for understanding why CCE \cite{pham2023circumventing} can attack E-CEMs and why STEREO \cite{srivatsan2025stereo} can robustly erase a target concept. We then discuss how to find effective reference subspaces for erasing narrow concepts (\cref{sec:refspace}), as well as the motivation for this subspace mapping approach, rather than simply pushing the target subspace away. Finally, we present our proposed method, the Subspace Mapping Approach for Concept Erasure (SuMa) (\cref{sec:suma}).

\subsection{Observations}\label{sec:analysis}
% We conducted a toy experiment using Concept-Ablation (CA) \cite{kumari2023ablating} to examine whether it could erase tokens that differ in form but convey the same concept. We began by applying Textual Inversion (TI) \cite{gal2022image} on SD1.4 \cite{RobinSD1.4} for 500 steps with 5 concepts reported in CA. For each concept we collect all learned tokens from steps 300 to 500 denoted as set $\mathcal{A}$. At this stage, all tokens were able to generate the target concept. We then trained CA to erase one randomly selected token from $\mathcal{A}$ and observed that the remaining tokens, even those differing in form, were also erased.
We conducted a toy experiment using Concept-Ablation (CA) \cite{kumari2023ablating} to investigate whether it can erase tokens that differ in form but convey the same underlying concept. We began by applying Textual Inversion (TI) on SD1.4 for 500 steps using five concepts reported in CA \cite{kumari2023ablating}. For each concept, we collected learned tokens from steps 300 to 500, denoted as set $\mathcal{A}$. At this stage, all tokens in $\mathcal{A}$ could generate the target concept. We then trained CA to erase one randomly selected token from $\mathcal{A}$ and observed that the remaining tokens, though differing in form, were also erased.

Next, we calculated the cosine similarity between the erased token and all tokens from $\mathcal{A}$, resulting in a distribution shown in Fig. \ref{fig:cosin-same-model}. We noted that the cosine similarity values ranged from \textbf{0.6} to \textbf{1.00}. We proceeded by running TI on the CA-based concept-erased model above. Upon reaching step 300, TI successfully recovered the erased concept. We then computed the cosine similarity between the new TI tokens and all tokens from $\mathcal{A}$ again and showed the result in Fig. \ref{fig:cosin-diff-model}. At this time, the cosine similarity ranged between \textbf{0.3} and \textbf{0.4}, significantly lower than that of the previously erased token. Based on these observations, we hypothesize that we can't represent the target concept as a single token but rather as a subspace; erasing a concept could be viewed as eliminating a subspace, and if the cosine similarity between this subspace and tokens that produce a similar concept exceeds \textbf{0.6}, those tokens are erased along with the target subspace.

To verify this hypothesis, we follow the procedure in STEREO \cite{srivatsan2025stereo} by iteratively erasing tokens found by TI using CA. The process is illustrated in Fig. \ref{fig: Stage1}. For each iteration, we collect the token from the final training step of TI. We aggregate tokens over $n$ iterations, denoted as $x_1, x_2, \dots, x_n \in \mathbb{R}^d$, ensuring that each token $x_i$ for $i \in [1, n]$ can generate the target concept. We then construct the matrix $\mathbf{S}_t = [x_1, \dots, x_t] \in \mathbb{R}^{d \times t}$ for $t \in [1, n-1]$. Next, we compute the angle between each token $x_j$ for $j \in [t+1, n]$ and the column space of $\mathbf{S}_t$, denoted by $c_{tj}$, using the formula:

\begin{equation} 
c_{tj} = \cos \left( x_j , \mathbf{S}_t \left( \mathbf{S}_t^T \mathbf{S}_t \right)^{-1} \mathbf{S}_t^T x_j \right),
\end{equation}
where $\mathbf{S}_t \left( \mathbf{S}_t^T \mathbf{S}_t^T \right)^{-1} \mathbf{S}_t^T$ is the projection matrix onto the column space of $\mathbf{S}_t$. We compute the average value of $c_{tj}$ for different values of $j$, denoted as $c_t$, and report them in Fig. \ref{fig:cosin-subspace}. Notably, when $t \geq 3$, $c_t$ exceeds \textbf{0.6}—the threshold we hypothesize is sufficient to cover the target concept. 
% implying that we only need maximum 3 iterations to collect sufficient set of TI tokens to completely remove the target concept. 
Then, we employ the second stage of STEREO to erase two sets of tokens: $\mathcal{S}_1 = \{x_1, x_2, x_3\}$ and $\mathcal{S}_2 = \{x_1, x_2\}$. We calculated the average Attack Success Rate (ASR) across all concepts, yielding $\mathcal{S}_1$ at $3.21\%$ and $\mathcal{S}_2$ at $85.32\%$. This indicates STEREO effectively defends against CCE when trained to erase $\mathcal{S}_1$, but not when only $\mathcal{S}_2$ is erased. An additional experiment on token similarity’s impact on erasure is detailed in Appendix \ref{sec:More_Toy_Exp}.

We found that the adversarial tokens identified by TI encompass those detected by UnlearnDiff. Specifically, we evaluated a model trained to erase $\mathcal{S}_1$ using UnlearnDiff, which yielded an average \textit{Attack Success Rate} of $1.52\%$, indicating robustness against UnlearnDiff. To explain this result, we provide an additional justification. Since UnlearnDiff identifies combinations of existing tokens, we leveraged the attention mechanism of the text encoder by using the $\langle \text{EOT} \rangle$ token to represent the target concept, and applied the same technique to all tokens from $\mathcal{S}_1$. After this processing, we computed a cosine similarity of \textbf{0.78} between the subspace formed by the $\langle \text{EOT} \rangle$ token from $\mathcal{S}_1$ and the one identified by UnlearnDiff, exceeding 0.6 and supporting our observations.

In summary, our analysis explains why STEREO can robustly remove the target concept. By applying TI on the concept-erased model iteratively in at least 3 iterations, it can collect a sufficient set of tokens spanning the target concept space. When erasing these tokens together, it successfully eradicates the model knowledge of the concept. 

\subsection{Reference Subspace Selection for Narrow Concepts }\label{sec:refspace}
CA \cite{kumari2023ablating} and ESD \cite{gandikota2023erasing} were the first methods proposed for concept erasure, but they present contradictory approaches. CA maps noise conditioned on the target concept to a more general representation, while ESD pushes the target noise as far as possible from the original. Studies \cite{lu2024mace, srivatsan2025stereo} show that CA outperforms ESD in terms of image quality. It is because mapping approach offers a clear direction to erase target concepts, whereas pushing approach can inadvertently affect unrelated concepts. The second stage of STEREO \cite{srivatsan2025stereo}, also a pushing approach, significantly degrades image quality. With this motivation, our paper proposes a mapping approach that erases tokens by mapping them to nearby non-target concept subspaces, preserving model utility. The key challenge is how to automatically select these reference subspaces. We also provide experimental results for Subspace Pushing, where the target subspace is simply pushed as far as possible from the original one, in Appendix \ref{sec:SubPush}.

We observe that constructing the reference subspace using tokens from the early training steps of TI is effective for narrow concepts, as early tokens typically converge to more general forms (e.g., grumpy cat to cat, or R2D2 robot to robot; see Section \ref{sec: Ablation Study} for details). Moreover, these tokens are closer to the target token learned from TI because they align along the same gradient direction. To validate this, we compare the Euclidean distance between the target token and early TI training tokens, as well as the target token and the textual token, as shown in Table \ref{tab:Distance}. The results indicate that while early tokens may represent concepts similar to the textual token, their distance from the target token is much smaller. This suggests that mapping the target token to an earlier token minimizes weight modifications, thereby better preserving the model's utility.

\begin{table}[t]
  \centering
  \begin{tabular}{@{}lcc@{}}
    \toprule
    Concept & TI Early Token & Textual Token \\ % Added a new column header
    \midrule
    \textit{R2D2 Robot} & 0.64 & 1.23 \\ 
    \textit{Garbage Truck} & 1.05 & 1.28 \\
    \textit{Grumpy Cat} & 0.98 & 1.22 \\
    \textit{English Springer Dog} & 0.72 & 0.96 \\
    \bottomrule
  \end{tabular}
  \vspace{-0.5em}
  \caption{The Euclid distance of the target token to the tokens from early TI training steps and the textual one.}
  \label{tab:Distance}
  \vspace{-1em} 
\end{table}

\begin{figure*}[t]
  \centering
  \begin{subfigure}[t]{0.49\linewidth} % Reduced width slightly for better spacing
    \includegraphics[width=\linewidth]{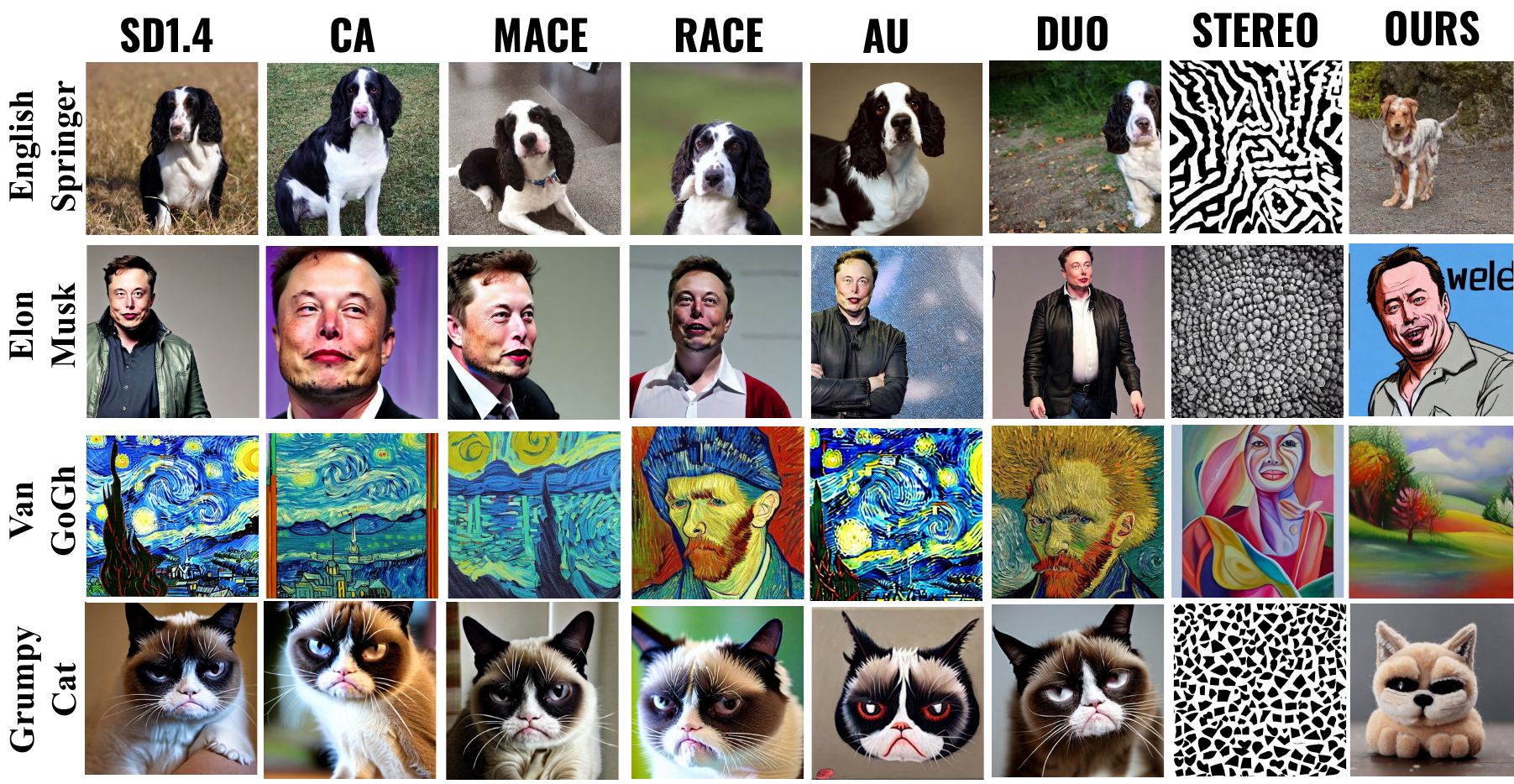} % Include Stage1 image
    % \caption{\textbf{Qualitative Comparison across 6 methods for the CCE attack.} We observe that both E-CEM methods, CA and RACE, are highly vulnerable to the CCE attack. STEREO’s erasure is overly aggressive, resulting in outputs that lack meaningful information except for Van Gogh-style elements; we notice that this concept is easier to erase than other concepts. Our method successfully defends against CCE by mapping the target concept to its reference concept.}
    \caption{\textbf{Qualitative Comparison across 6 methods for the CCE attack.} Both E-CEM methods, CA and MACE, are vulnerable to the CCE attack. STEREO's erasure is too aggressive, leaving only Van Gogh-style elements, with this concept being easier to erase. Our method defends against CCE by mapping the target concept to its reference concept.}
    \label{fig: Qua1}
  \end{subfigure}%
  \hspace{0.01\linewidth} 
  \begin{subfigure}[t]{0.49\linewidth}
    \includegraphics[width=\linewidth]{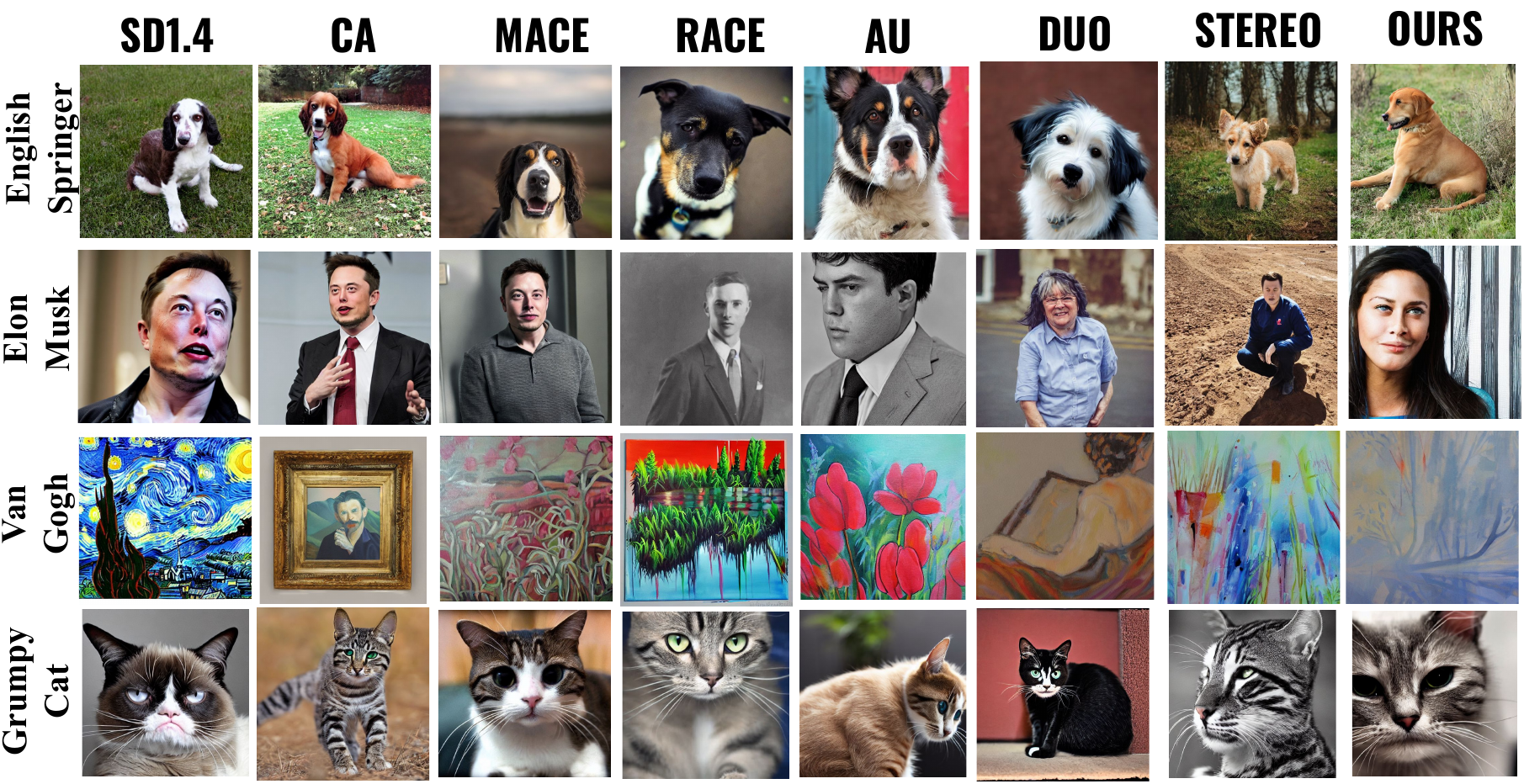} % Include Stage2 image
    % \caption{\textbf{Qualitative Comparison across 6 methods for image generation with the textual target term.} We observe that CA does not perform well for "English Springer" and "Elon Musk." MACE performs effectively for these concepts, maintaining good image quality. RACE shows a slight decrease in image quality, while STEREO significantly impacts image quality. Our method appears to be in between E-CEMs and C-CEMs in terms of image quality.}
    % \caption{\textbf{Qualitative Comparison across 6 methods for image generation with the textual target term.} We observe that CA struggles with "English Springer" and "Elon Musk." MACE performs well, preserving image quality. RACE shows a slight decrease, while STEREO significantly reduces image quality. Our method balances between E-CEMs and C-CEMs in image quality.}

    \caption{\textbf{Qualitative Comparison across 6 methods for image generation.} CA struggles with "English Springer" and "Elon Musk." MACE performs well, preserving image quality. RACE shows slight degradation, and STEREO significantly reduces quality. Our method strikes a balance between E-CEMs and C-CEMs in terms of quality.}

    \label{fig: Qua2}
  \end{subfigure}
  \vspace{-2mm}
  \caption{\textbf{Qualitative Results.} We present qualitative results comparing five methods against the CCE attack and textual erasure.}
    \vspace{-2mm}
  \label{fig:Qualiative}
\end{figure*}

\subsection{Proposed Method}\label{sec:suma}

Our framework, \textit{SuMa}, consists of two stages, as illustrated in Fig. \ref{fig:Main_Framework}. The first stage, termed \textbf{Subspace Construction}, aims to identify Textual Inversion tokens to establish both sets of Target Subspace $\{\mathcal{U}^i\}$ and Reference Subspace $\{\mathcal{V}^i\}$. Here, $\mathcal{U}^i$ represents $i$-th subspace (corresponding to $i$-th Unet layer) of the concept to be erased, while $\mathcal{V}^i$ is $i$-th the subspace into which we intend to map our $i$-th target subspace. In the second stage, \textbf{Subspace Elimination}, we refine the to-$K$ and to-$V$ weights of the Cross-Attention Layer in U-Net, denoted as $W_k$ and $W_v$, allowing us to project the basis of  $\mathcal{U}^i$ onto $\mathcal{V}^i$. Due to the identical functionality of $W_k$ and $W_v$ in our work, we denote both as $W_k$ for simplicity and readability.

\minisection{Subspace Construction.} Starting with the original model, denoted as $\theta$, and a concept to be removed, $c$, we apply a Concept Ablation ($\mathcal{M}$) to eliminate $c$ from $\theta$, resulting in a modified model $\theta^{\prime}$. This process is repeated over $l$ steps. At $j$-th iteration, we apply Textual Inversion (TI) for the concept to be erased from $\theta^{\prime}$, yielding two tokens: one from the final step of TI training, denoted as $\langle u \rangle_j$, and another from an earlier step, denoted as $\langle v \rangle_j$. Subsequently, we construct prompts in the formats \textit{"A photo of $\langle u\rangle_j$"} and \textit{"A photo of $\langle v \rangle_j$"}, which are then passed through the Text Encoder $\mathcal{T}$ to extract embeddings for the tokens $\langle u_j \rangle$ and $\langle v_j \rangle$. We then apply the transformation $W_k^i$, where $i$ represent $i$-th layer in Unet, to these extracted embeddings, treating the outputs as $\langle u^{\prime} \rangle_j^{i}$ and $\langle v^{\prime} \rangle_{j}^i$ as basis vectors for the subspaces 
 $\mathcal{U}^i$ and $\mathcal{V}^i$ respestively. Formally, this process is defined as follows:
\begin{equation}
\begin{split}
    \langle u^{\prime} \rangle_j^i &= W_k^i \mathcal{T}(p + \langle u \rangle_j), \\
    \langle v^{\prime} \rangle_j^i &= W_k^i \mathcal{T}(p + \langle v \rangle_j),
\end{split}
\end{equation}
where $p$ represents the prompt template. Finally, we fine-tune $\theta^{\prime}$ using $\mathcal{M}$ to remove $\langle u_j \rangle$, resulting in an updated model $\theta^{\prime}$ at the end of each iteration (Fig. \ref{fig: Stage1}).

\minisection{Subspace Elimination.} Fig. \ref{fig: Stage2} illustrates this process. After constructing sets of the subspaces $\{\mathcal{U}^i\}_{i=1}^M$ and $\{\mathcal{V}^i\}_{i=1}^M$, with $M$ is number of layer in Unet, our goal is to eliminate all $\mathcal{U}^i$  by fine-tuning $W_k$ so that the basis vectors of $\mathcal{U}^i$ are projected onto $\mathcal{V}^i$. %We define $\mathcal{U}$ as the span of the basis vectors $\langle u^{\prime} \rangle_1, \langle u^{\prime} \rangle_2, \dots, \langle u^{\prime} \rangle_l$, and similarly, $\mathcal{V}$ as the span of the basis vectors $\langle v^{\prime} \rangle_1, \langle v^{\prime} \rangle_2, \dots, \langle v^{\prime} \rangle_l$. 
To construct the projection matrix onto $\mathcal{V}^i$, we first define the matrix $\mathbf{V}^i = [\langle v^{\prime} \rangle_1^i, \langle v^{\prime} \rangle_2^i, \dots, \langle v^{\prime} \rangle_l^i] \in \mathbb{R}^{d \times l}$, with $d$ is the dimension of token embedding vector, and the projection matrix $\mathbf{P}_{\mathbf{V}^i}$ can then be formulated as:

\begin{equation}
    \mathbf{P}_{\mathbf{V}^i} = \mathbf{V}^i \left( \left( \mathbf{V}^i \right)^T \mathbf{V}^i \right)^{-1} \left( \mathbf{V}^i \right)^T
    \label{eq:Proj_matrix}
\end{equation}
We define \( L_{proj} \) as the objective to project all basis of \(\mathcal{U}^i\) onto \(\mathcal{V}^i\) by following formula:
\begin{equation}
    L_{proj} = \sum_{i=1}^M\sum_{j=1}^l \left\| W_k^i \mathcal{T} \left (p + \langle u \rangle_j \right) - \mathbf{P}_{\mathbf{V}^i} W_k^i \mathcal{T} \left( p +\langle u \rangle_j \right) \right\|_2^2
    \label{loss: distance}
\end{equation}
As per \cite{bui2024erasing}, we add a regulization term to keep the \textit{Reference Subspace} \(\mathcal{V}^i\) unchanged during fine-tuning:

\begin{equation}
    L_{reg} = \sum_{i=1}^M\sum_{j=1}^l \left\| W_k^i \mathcal{T} \left (p + \langle v \rangle_j \right) - \mathbf{P}_{\mathbf{V}^i} W_k^i \mathcal{T} \left( p +\langle v \rangle_j \right) \right\|_2^2
    \label{loss: reg}
\end{equation}
The objective function for subspace mapping, \( L_{sub} \), is then defined as the weighted sum of these two terms:
\begin{equation}
    L_{sub} = L_{proj} + \lambda_{reg}L_{reg},
\end{equation}
% where $\lambda_{reg}$ is a hyper-parameter with the default value as 1. If we use \( L_{sub} \) to fine-tune \( \theta \) and obtain \( \theta^{\prime} \), then apply CCE on \( \theta^{\prime} \) to find a new target token, this token will still be able to generate the target concept when conditioned on \( \theta^{\prime} \), but it did not generate the concept when conditioned on \( \theta \) (We provide more details in the Appendix \ref{sec:ModifyCrossAttn}). Our assumption here is that modifying only the cross-attention layer works fine for erasing a single token, as shown in \cite{lu2024mace}, but when erasing a set of tokens, it affects the SDE path, leading to this phenomenon. To address this, we use the loss function proposed in CA to reduce the impact on the SDE path, \textbf{we discuss the reason why we chose CA in the Appendix}. In short, their objective can be formulated as:
where $\lambda_{reg}$ is a hyperparameter (default 1). If we use \( L_{sub} \) to fine-tune \( \theta \) to obtain \( \theta^{\prime} \) and use CCE to find a new target token, this token still generates the target concept when conditioned on \( \theta^{\prime} \), but not on \( \theta \) (see Appendix \ref{sec:ModifyCrossAttn}). We assume modifying only the cross-attention layer is effective for erasing a token, as in \cite{lu2024mace}, but affects the SDE trajectory when erasing multiple tokens. To mitigate this, we use the CA loss function and discuss our choice in Section \ref{sec: Ablation Study}.
. In short, their objective is:
\begin{equation}
    L_{\text{CA}} = \mathbb{E}_{z_t, c, t} \left[ \| \Phi^{\prime}(z_t, \mathcal{T}(p + c'), t) - \Phi(z_t, \mathcal{T}(p + c), t) \|_2^2 \right]
\end{equation}
where $\Phi^{\prime}$ is the model to learn, $\Phi$ is pretrained, $\mathcal{T}$ is Text-Encoder, $p$ is prompt template, $c$ is target concept, and $c'$ is its general term (e.g., \textit{grumpy cat} vs. \textit{cat}). Our objective is:

\begin{equation} 
    W^*_k = \operatorname*{argmin}_{W_k} \left( L_{\text{final}} \right) = \operatorname*{argmin}_{W_k} 
    \left( L_{\text{CA}} + L_{\text{sub}} \right).
\label{equal:final}
\end{equation}

%% file: sec/4_Experiment.tex
\section{Experiments}

\begin{table*}[t]
    \centering
    \small % Keep font size small for readability
    \setlength{\tabcolsep}{2.5pt} 
    \begin{tabular}{cc}
        % First Column: Subclass and Identity
        \scalebox{.92}{\begin{minipage}{0.48\textwidth}
            \centering
            \begin{tabular}{l|c|c|c|c}
                \toprule
                \textbf{Method} & \textbf{CCE $\downarrow$} & \textbf{UD $\downarrow$} & \textbf{FID $\downarrow$} & \textbf{CLIP $\uparrow$} \\
                \midrule
                \multicolumn{5}{c}{\textbf{Subclass}} \\
                \midrule
                SD \cite{RobinSD1.4} & \failed{0.84 / \textit{0.83}} & \failed{0.74 / \textit{0.76}} & 17.04 / \textit{16.95} & 0.33 / \textit{0.32} \\
                CA \cite{kumari2023ablating} & \failed{0.78 / \textit{0.75}} & \failed{0.67 / \textit{0.62}} & 19.28 / \textit{19.26} & 0.32 / \textit{0.32} \\
                MACE \cite{lu2024mace} & \failed{0.86 / \textit{0.84}} & \failed{0.73 / \textit{0.68}} & 17.15 / \textit{16.98} & 0.29 / \textit{0.31} \\
                RACE \cite{kim2024race} & \failed{0.74 / \textit{0.73}} & \success{0.17 / \textit{0.18}} & 26.34 / \textit{26.22} & 0.27 / \textit{0.28} \\
                AdvUnlearn \cite{zhang2024defensive} & \failed{0.76 / \textit{0.74}} & \success{0.18 / \textit{0.16}} & 18.47 / \textit{18.39} & 0.30 / \textit{0.31} \\
                DUO \cite{parkdirect} & \failed{0.65 / \textit{0.63}} & \failed{0.62 / \textit{0.65}} & 17.09 / \textit{16.99} & 0.29 / \textit{0.30} \\
                STEREO \cite{srivatsan2025stereo} & \success{0.05 / \textit{0.03}} & \success{0.05 / \textit{0.04}} & 27.33 / \textit{27.20} & 0.30 / \textit{0.28} \\
                \textbf{Ours} & \success{0.12 / \textit{0.10}} & \success{0.08 / \textit{0.06}} & 18.33 / \textit{18.15} & 0.31 / \textit{0.30} \\
                \midrule

                \multicolumn{5}{c}{\textbf{Identity}} \\
                \midrule
                SD \cite{RobinSD1.4} & \failed{0.91 / \textit{0.90}} & \failed{0.86 / \textit{0.84}} & 17.04 / \textit{16.95} & 0.33 / \textit{0.32} \\
                CA \cite{kumari2023ablating} & \failed{0.87 / \textit{0.89}} & \failed{0.79 / \textit{0.76}} & 18.21 / \textit{18.27} & 0.31 / \textit{0.30} \\
                MACE \cite{lu2024mace} & \failed{0.89 / \textit{0.87}} & \failed{0.65 / \textit{0.63}} & 16.74 / \textit{16.69} & 0.29 / \textit{0.28} \\
                RACE \cite{kim2024race} & \failed{0.78 / \textit{0.79}} & \failed{0.47 / \textit{0.46}}& 24.32 / \textit{24.17} & 0.27 / \textit{0.29} \\
                AdvUnlearn \cite{zhang2024defensive} & \failed{0.77 / \textit{0.69}} & \failed{0.52 / \textit{0.53}} & 17.53 / \textit{17.69} & 0.31 / \textit{0.31} \\
                DUO \cite{parkdirect} & \failed{0.73 / \textit{0.71}} & \failed{0.67 / \textit{0.73}} & 17.32 / \textit{17.11} & 0.29 / \textit{0.30} \\
                STEREO \cite{srivatsan2025stereo} & \success{0.02 / \textit{0.01}} & \success{0.03 / \textit{0.01}}& 26.33 / \textit{26.15} & 0.29 / \textit{0.28} \\
                \textbf{Ours} & \success{0.09 / \textit{0.07}} & \success{0.18 / \textit{0.14}} & 17.94 / \textit{17.89} & 0.31 / \textit{0.29} \\
                \bottomrule
            \end{tabular}
        \end{minipage}}
        \hspace{1.5em} % Add horizontal space between the two minipages to prevent overlap
        &
        % Second Column: Artistic and Instance
        \scalebox{.92}{\begin{minipage}{0.48\textwidth}
            \centering
            \begin{tabular}{lc|c|c|c}
                \toprule
                \textbf{Method} & \textbf{CCE $\downarrow$} & \textbf{UD $\downarrow$} & \textbf{FID $\downarrow$} & \textbf{CLIP $\uparrow$} \\
                \midrule
                \multicolumn{5}{c}{\textbf{Artistic}} \\
                \midrule
                SD \cite{RobinSD1.4} & \failed{0.86 / \textit{0.83}} & \failed{0.82 / \textit{0.85}} & 17.04 / \textit{16.95} & 0.33 / \textit{0.32} \\
                CA \cite{kumari2023ablating} & \failed{0.79 / \textit{0.78}} & \failed{0.74 / \textit{0.76}} & 18.02 / \textit{17.85} & 0.34 / \textit{0.31} \\
                MACE \cite{lu2024mace} & \failed{0.84 / \textit{0.82}} & \failed{0.79 / \textit{0.83}} & 16.57 / \textit{16.58} & 0.32 / \textit{0.33} \\
                RACE \cite{kim2024race} & \failed{0.82 / \textit{0.79}} & \success{0.06 / \textit{0.04}} & 21.32 / \textit{21.10} & 0.30 / \textit{0.31} \\
                AdvUnlearn \cite{zhang2024defensive} & \failed{0.83 / \textit{0.76}} & \success{0.03 / \textit{0.05}}& 17.73 / \textit{17.60} & 0.32 / \textit{0.33} \\
                DUO \cite{parkdirect} & \failed{0.74 / \textit{0.72}} & \failed{0.67 / \textit{0.73}} & 17.74 / \textit{17.89} & 0.30 / \textit{0.31} \\
                STEREO \cite{srivatsan2025stereo} & \success{0.00 / \textit{0.02}} & \success{0.00 / \textit{0.01}} & 28.04 / \textit{27.90} & 0.30 / \textit{0.31} \\
                \textbf{Ours} & \success{0.05 / \textit{0.04}} & \success{0.07 / \textit{0.05}} & 17.18 / \textit{17.21} & 0.33 / \textit{0.31} \\
                \midrule
                \multicolumn{5}{c}{\textbf{Instance}} \\
                \midrule
                SD \cite{RobinSD1.4} & \failed{0.97 / \textit{0.94}} & \failed{0.94 / \textit{0.93}} & 17.04 / \textit{16.95} & 0.33 / \textit{0.32} \\
                CA \cite{kumari2023ablating} & \failed{0.97 / \textit{0.93}} & \failed{0.92 / \textit{0.89}} & 18.43 / \textit{18.30} & 0.30 / \textit{0.31} \\
                MACE \cite{lu2024mace} & \failed{0.97 / \textit{0.92}} & \failed{0.93 / \textit{0.89}} & 16.93 / \textit{16.80} & 0.30 / \textit{0.31} \\
                RACE \cite{kim2024race} & \failed{0.95 / \textit{0.92}} & \success{0.06 / \textit{0.07}} & 24.32 / \textit{24.15} & 0.28 / \textit{0.29} \\
                AdvUnlearn \cite{zhang2024defensive} & \failed{0.95 / \textit{0.93}} & \success{0.03 / \textit{0.04}} & 18.37 / \textit{18.20} & 0.32 / \textit{0.33} \\
                DUO \cite{parkdirect} & \failed{0.92 / \textit{0.91}} & \failed{0.89 / \textit{0.87}} & 18.31 / \textit{18.92} & 0.30 / \textit{0.31} \\
                STEREO \cite{srivatsan2025stereo} & \success{0.03 / \textit{0.02}} & \success{0.05 / \textit{0.03}} & 27.04 / \textit{26.90} & 0.31 / \textit{0.32} \\
                \textbf{Ours} & \success{0.17 / \textit{0.15}} & \success{0.19 / \textit{0.16}} & 22.34 / \textit{22.20} & 0.29 / \textit{0.30} \\
                \bottomrule
            \end{tabular}
        \end{minipage}}
    \end{tabular}
    % \caption{\textbf{Evaluation of Erasing across 4 Categories.} We average the results of all target concepts within each category across three metrics and compare our framework, SuMa, with current state-of-the-art methods. With ASRs, we highlight the successful attacks (ASR $\ge$ 20\%) in \failed{red} and the defeated attacks (ASR $<$ 20\%) in \success{green}. The data in each cell represent the result of erasing a concept from the pretrained SD1.4 / \textit{SD1.5}.}
    \vspace{-2mm}
    \caption{\textbf{Evaluation of Erasing across 4 Categories.} We average the results of all target concepts in each category across three metrics and compare our framework, SuMa, with state-of-the-art methods. For ASRs, successful attacks (ASR $\ge$ 20\%) are in \failed{red}, and defeated attacks (ASR $<$ 20\%) in \success{green}. The data in each cell show the result of erasing a concept from pretrained SD1.4 / \textit{SD1.5}.}
    \vspace{-2mm}
    \label{tab: Main Result}
\end{table*}

\subsection{Experimental Setting} \label{sec: Exp Setting}

\minisection{Dataset.} We evaluate SuMa across four tasks: subclass erasure, celebrity erasure, artistic style erasure, and instance erasure. For \textit{\textbf{subclass erasure}}, we focus on a subset of ImageNette \cite{howard2020fastai} known for its distinctive features. Three subclasses are selected for the test, including \textit{English Springer Dog}, \textit{Garbage Truck}, and \textit{Golf Ball}. For \textit{\textbf{celebrity erasure}}, we select 3 celebrities in the Celebrity-1000 dataset \cite{Tonyassi}, including  \textit{David Beckham}, \textit{Elon Musk}, and \textit{Adam Lambert} . For \textit{\textbf{artistic style erasure}}, following the approach in \cite{gandikota2023erasing, kim2024race, srivatsan2025stereo}, we focus on the style of \textit{Van Gogh}, which is both famous and straightforward to verify in outputs. For instance erasure, we conduct experiments on three copyrighted concepts: \textit{Grumpy Cat}, \textit{Mickey Mouse}, and the \textit{R2D2 robot}. Note that our method is specifically designed to address the problem with narrow concepts, so it does not work as well for NSFW concepts. However, as mentioned earlier, our work is complementary to DUO, so it can be easily integrated with this method to completely erase all kinds of concepts, demonstrated in the Appendix \ref{sec:NSFW_Concepts}.

% \minisection{Baseline}. We compare SuMa with four baselines, including two E-CEMs (CA \cite{kumari2023ablating} and MACE \cite{lu2024mace}) and four C-CEMs (RACE \cite{kim2024race}, AdvUnlearn (AU) \cite{zhang2024defensive}, STEREO \cite{srivatsan2024stereo}, DUO \cite{parkdirect}). For CA, MACE, RACE, and AU, we use their publicly available code and recommended settings to erase our chosen concepts. For STEREO and DUO, due to their code not being publicly available, we re-implement them with the settings recommended in their papers and test the implementation to ensure we obtain consistent results as reported.
\minisection{Baseline}. We compare SuMa with four baselines: two E-CEMs (CA \cite{kumari2023ablating} and MACE \cite{lu2024mace}) and four C-CEMs (RACE \cite{kim2024race}, AdvUnlearn (AU) \cite{zhang2024defensive}, STEREO \cite{srivatsan2025stereo}, DUO \cite{parkdirect}). For CA, MACE, RACE, and AU, we use their public code and recommended settings to erase concepts. For STEREO and DUO, we re-implement their methods using paper recommendations and verify consistency with reported results.

\minisection{Circumventing Methods.} We evaluate the robustness of our method under CCE \cite{pham2023circumventing} and UnlearnDiff (UD) \cite{zhang2025generate}. For CCE, we use a different set of images from the \textbf{Subspace Construction} stage to ensure the model didn't overfit to this image set. For both attack methods, we use publicly available code to generate adversarial prompts.
 
\minisection{Evaluation Metrics.} 
% Similar to previous techniques, our erased models do not generate the target concept when using the same prompt; we include this trivial test in the Supplementary. Here, we focus on more challenging experiments to verify the completeness and effectiveness of the proposed algorithm. 
First, to measure \textbf{\textit{effectiveness}}, for each erased model, we generate 5,000 images using prompts from the MS-COCO validation set. We then evaluate the model quality when generating such non-target prompts using \textbf{CLIP} scores (higher is better) and \textbf{FID} scores (lower is better). To measure \textbf{\textit{completeness}}, in the \textbf{CCE} attack, for each erased model, we execute the circumventing method to acquire the adversarial token \( c_{adv} \). We then generate 200 images using a template prompt: \textit{``A photo of a \{$c_{adv}$\}}." For style erasure, we use the template \textit{``A painting in the style of \{$c_{adv}$\}"}. In the \textbf{UD} attack, we execute its code on our erased model to get the adversarial prompt and pass the prompt to the erased model. We report the attack success rate (ASR), where a lower score indicates better performance, following the method in \cite{srivatsan2025stereo, kim2024race}. For subclass erasure tasks, we utilize \textbf{ResNet-50} \cite{he2016deep}, pretrained on \textbf{ImageNet} \cite{russakovsky2015imagenet}, to measure ASR. For celebrity detection, we use the \textbf{GIPHY Celebrity Detector (GCD)} \cite{GcD} as the classifier. To evaluate artistic style, we follow the approach in \cite{kim2024race} by fine-tuning a \textbf{ViT-based} model \cite{alexey2020image} on the \textbf{WikiArt} dataset \cite{saleh2015large}. Finally, for instance detection, due to the lack of pretrained models or datasets, we use \textbf{LLAVA-7B} \cite{liu2024visual} with a template prompt: \textit{``Does this image contain \{object\}? Answer Yes or No."} for evaluation.

\begin{figure}[t]
  \centering
   \includegraphics[width=.8\linewidth]{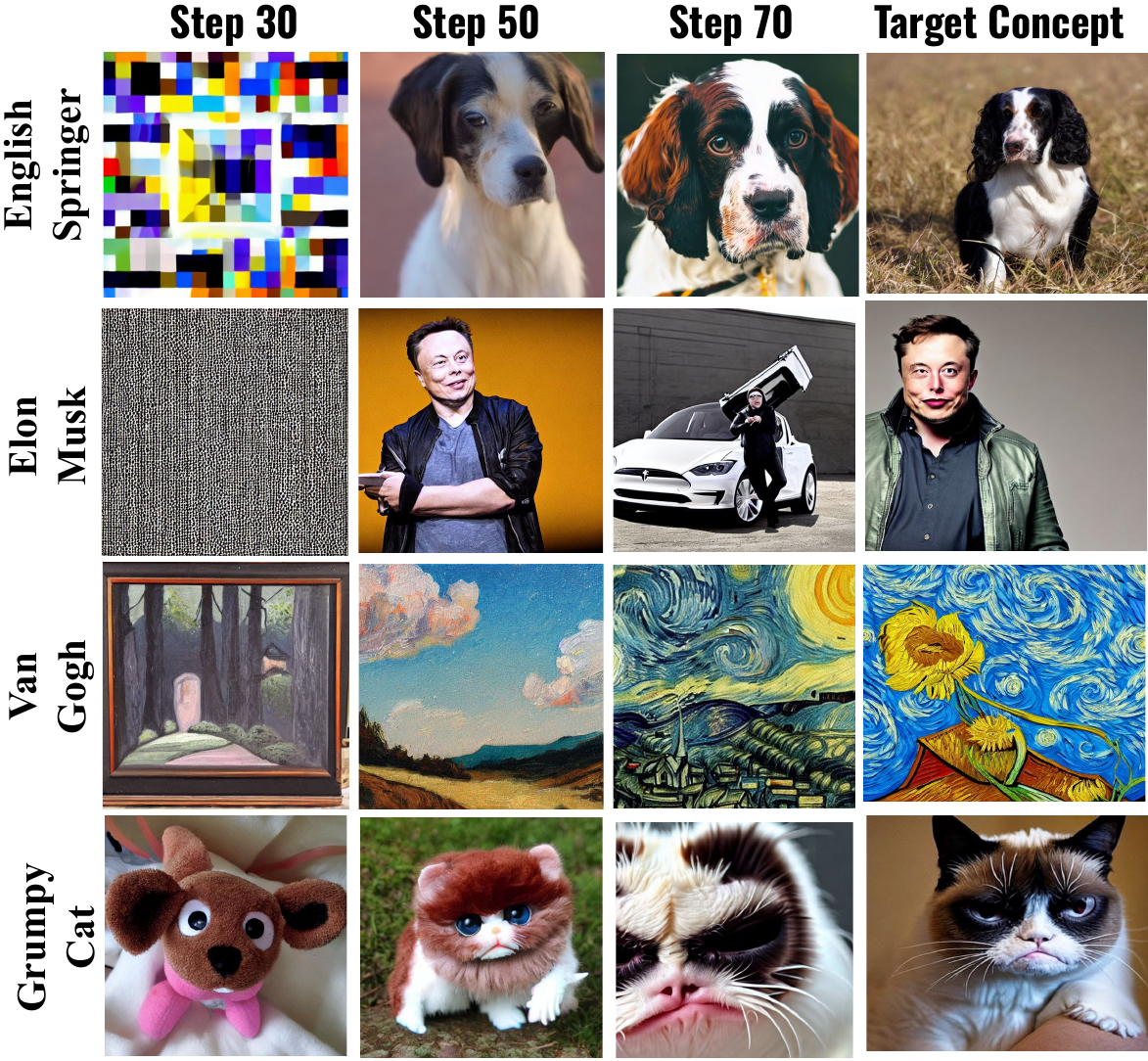}
   \vspace{-2mm}
   \caption{Visualization of Textual Inversion tokens at different numbers of TI training steps.} 
   \label{fig:Ablation1}
   \vspace{-1.65em}
\end{figure}

\minisection{Implementation Details.} We use SD1.4 and SD1.5 as the pretrained model. In all experiments, we fine-tune all cross-attention layers. In the \textbf{Subspace Construction} stage, we use the same setting as CA \cite{kumari2023ablating} for concept erasion. Training TI is set for 500 steps, with the early step to get token $\langle v \rangle$ set to 50, the initial token is \textit{toy}, with Adam optimization with learning rate, is 2e-6. For the \textbf{Subspace Elimination} stage, we fine-tune most objects for 500 steps, except for subclass objects, which require 750 steps to ensure model safety. We use the Adam optimizer with a learning rate of 2e-5. All experiments are done on 2 NVIDIA A100 GPUs.

\begin{table*}[t]
    \centering

    \setlength{\tabcolsep}{4pt}
    \small
    \begin{tabular}{lcccccccccccc}
        \toprule
        \textbf{\textit{Reference Subspace}} & \multicolumn{3}{c}{\textbf{Subclass}} & \multicolumn{3}{c}{\textbf{Identity}} & \multicolumn{3}{c}{\textbf{Artistic}} & \multicolumn{3}{c}{\textbf{Instance}} \\
        \cmidrule(lr){2-4} \cmidrule(lr){5-7} \cmidrule(lr){8-10} \cmidrule(lr){11-13}
        & ASR $\downarrow$ & FID $\downarrow$ & CLIP $\uparrow$ & ASR $\downarrow$ & FID $\downarrow$  & CLIP $\uparrow$ & ASR $\downarrow$ & FID $\downarrow$  & CLIP $\uparrow$ & ASR $\downarrow$ & FID $\downarrow$  & CLIP $\uparrow$ \\
        \midrule
        Step-30 &  0.09  & 19.11 & 0.30 & 0.08 & 18.34 & 0.29 & 0.05 & 18.82 & 0.31 & 0.17 & 23.96  & 0.27 \\
        Step-50 &0.12 & 18.33 & 0.31 & 0.09 &  17.94  & 0.31 & 0.05 & 17.18 & 0.33 & 0.17 & 22.34 & 0.29 \\
        Step-70& 0.56 & 17.32 & 0.33 & 0.67 & 17.92 & 0.32 & 0.37 & 17.05 & 0.33 & 0.83 & 17.74 & 0.32  \\
        \midrule
        General Term & 0.18 & 18.27 & 0.32 & 0.17 & 18.02 & 0.32 & 0.11 & 18.21 & 0.33 & 0.15 & 23.34 & 0.27 \\
        \bottomrule
    \end{tabular}
    \vspace{-2mm}
    \caption{\textbf{Ablation Study for the selection of \textit{Reference subspace.}}}
    \vspace{-4mm}
    \label{tab: Ablation Reference Subspace}
\end{table*}

\subsection{Results}

\minisection{Quantitative Results.} We average each metric from Section \ref{sec: Exp Setting} for each category and report the results in Table \ref{tab: Main Result}. For all baselines, we ensure the erased model works well for erasing the textual prompt, and we report results for this trivial task in the Appendix \ref{sec:StandartTest}. For \textbf{completeness} evaluation, our method achieves an ASR score below 0.2 for all categories under both \textbf{CCE} and \textbf{UD} attacks. Compared to \textbf{STEREO}, our method has a slightly higher ASR score (0.02 to 0.14 higher), but within an acceptable range. \textbf{RACE} and \textbf{AdvUnlearn} show safety against the \textbf{UD} attack in the Subclass, Artistic, and Instance categories but fail under Identity. This is due to the subtle facial features of human identity, which make this category more fine-grained and challenging. However, both fail under the \textbf{CCE} attack, showing that protecting the model against \textbf{CCE} is harder than \textbf{UD}. The main reason is that \textbf{RACE} and \textbf{AdvUnlearn} erase adversarial tokens based on \textbf{UnlearnDiff}, but these tokens don’t fully represent the target concept, protecting against \textbf{UD} but not \textbf{CCE}. \textbf{DUO} works well for \textbf{NSFW} concepts (results in Appendix \ref{sec:NSFW_Concepts}) but fails for narrow concepts. For \textbf{effectiveness}, although \textbf{STEREO} performs well under both \textbf{CCE} and \textbf{UD}, image quality is significantly affected, resulting in an \textbf{FID} score roughly double that of other \textbf{E-CEMs}, making it less effective. Our method, while not achieving a perfect ASR, achieves an \textbf{FID} score comparable to other \textbf{E-CEMs} in the Subclass, Artistic, and Identity categories, outperforming \textbf{CA} with an \textbf{FID} around 0.85 lower. We found that Instance category is harder to be erased, so we fine-tuned it for 750 steps, resulting in a higher \textbf{FID} compared to the others, fine-tuned for 500 steps.

\begin{figure}[t]
  \centering
   \includegraphics[width=.98\linewidth]{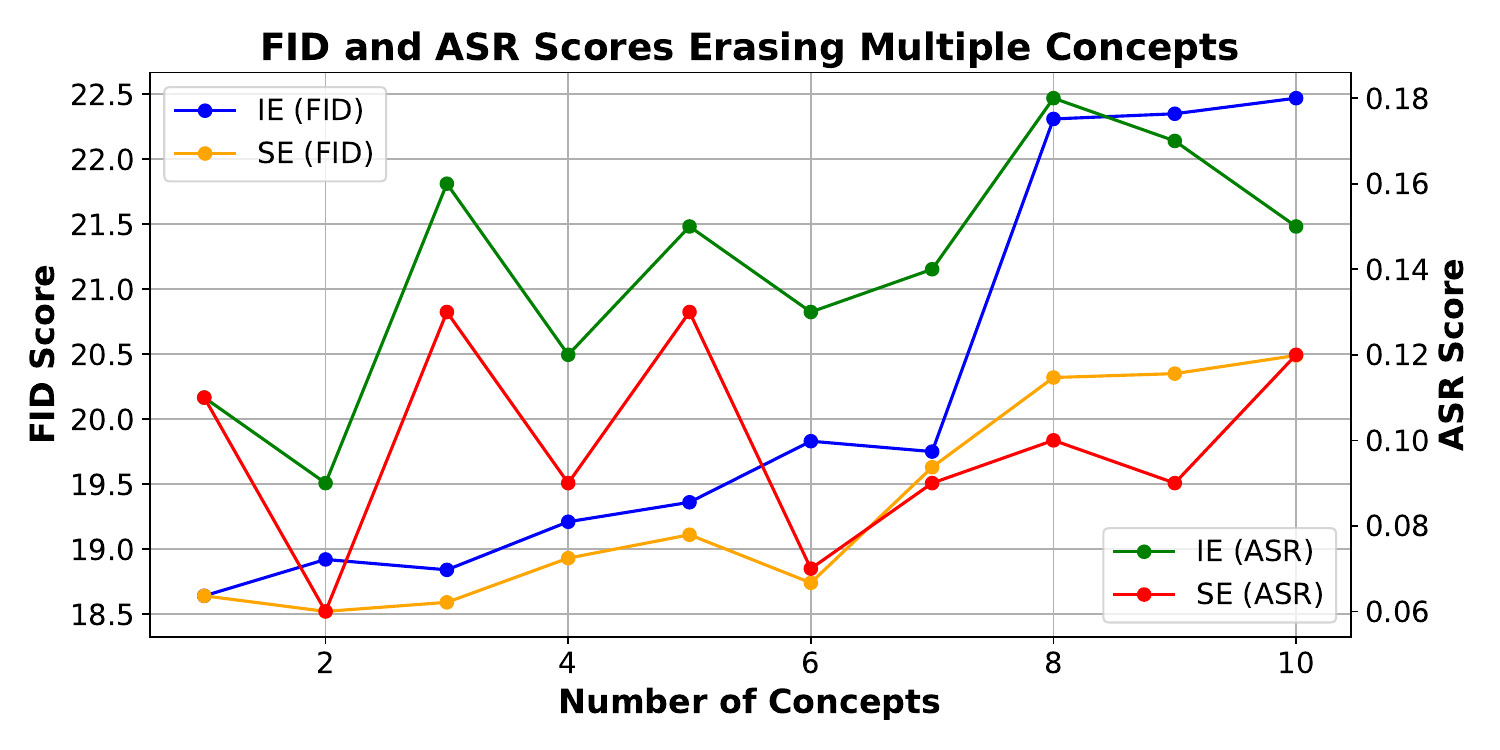}
   \vspace{-4mm}
   \caption{Multiple Concept Erasure Result.} 
   \label{fig:Multi concpets}
   \vspace{-4mm}
\end{figure}

% \minisection{Qualitative Result.} We provide some qualitative results across these methods in Fig. \ref{fig:Qualiative}. Fig. \ref{fig: Qua1} shows some results of these models against the CCE attack, \textbf{we include qualitative result for UD in the Appendix } . Here we can see that CA, MACE, RACE, and AdvUnlearn are highly vulnerable to this attack. STEREO completely removes these concepts; its CCE attack returns nothing but at the cost of image quality (see the 6th column in Fig. \ref{fig: Qua2}). Our method successfully converts ``Springer Dog" to ``Dog", ``Elon Musk" to a general person, and ``Grumpy Cat" to "Cat toy", while maintaining image quality comparable to other E-CEM methods (last column in Fig. \ref{fig: Qua2}).

\begin{table}[t]
    \centering
    \setlength{\tabcolsep}{7pt} % Further reduce space between columns
    \scriptsize % Use a smaller font size
    \renewcommand{\arraystretch}{1.1} % Adjust row height slightly for readability
    \begin{tabular}{lcccc}
        \toprule
        \textbf{\textit{Regulization}} & \textbf{Subclass} & \textbf{Identity} & \textbf{Artistic} & \textbf{Instance} \\
        \midrule
        \multirow{3}{*}{\centering \xmark } 
        & ASR: 0.09 & ASR: 0.06 & ASR: 0.05 & ASR: 0.15 \\
        & FID: 19.21 & FID: 20.31 & FID: 19.32 & FID: 24.57 \\
        & CLIP: 0.29 & CLIP: 0.29 & CLIP: 0.30 & CLIP: 0.26 \\
        
        \midrule
        \multirow{3}{*}{\ding{51}} 
        & ASR: 0.12 & ASR: 0.09 & ASR: 0.05 & ASR: 0.17 \\
        & FID: 18.33 & FID: 17.94 & FID: 17.18 & FID: 22.34 \\
        & CLIP: 0.31 & CLIP: 0.31 & CLIP: 0.33 & CLIP: 0.29 \\
        
        \bottomrule
    \end{tabular}
    \vspace{-2mm}
    \caption{\textbf{Ablation Study for the need of Subspace Regulization.}}
    \vspace{-3mm}
    \label{tab:Ablation Reg Subspace}
\end{table}

\minisection{Qualitative Results} are provided in Fig. \ref{fig:Qualiative}. Fig. \ref{fig: Qua1} shows results of these models against the CCE attack, and we include qualitative results for UD in the Appendix \ref{sec:Additional Qualiatative}. CA, MACE, RACE, and AdvUnlearn are highly vulnerable to this attack. STEREO completely removes these concepts, but at the cost of image quality ($6^{th}$ column in Fig. \ref{fig: Qua2}). Our method successfully converts ``Springer Dog" to ``Dog," ``Elon Musk" to a general person, and ``Grumpy Cat" to ``Cat toy," while maintaining image quality comparable to other E-CEM methods (last column in Fig. \ref{fig: Qua2}).

% \minisection{Multiple Concepts Erasure.} We provide the results of erasing multiple concepts in Fig. \ref{fig:Multi concpets}. Here, we test the ASR with the \textbf{CCE} attack and evaluate two settings: Iterative Erase (IE) and Simultaneous Erase (SE) of concepts. \textbf{IE} means that after we erase one concept, we continue to fine-tune the weights to erase other concepts. \textbf{SE} means that we sum all \( L_{sub} \) to erase all concepts in one fine-tuning, with the number of fine-tuning steps set to 1000 for this experiment. With both settings, our method is still able to defend against \textbf{CCE} and maintain the model's utility. We note that with \textbf{SE}, we achieve better results in terms of FID score. This is because summing all \( L_{sub} \) means we preserve a broader range of knowledge from \( L_{reg} \).

\minisection{Multiple Concepts Erasure.} We provide results for erasing multiple concepts in Fig. \ref{fig:Multi concpets}. We test the ASR with the \textbf{CCE} attack and evaluate two settings: Iterative Erase (IE) and Simultaneous Erase (SE). \textbf{IE} means we fine-tune the weights after erasing one concept to erase others. \textbf{SE} sums all \( L_{sub} \) to erase all concepts in one fine-tuning, with 1000 steps fine-tuning for this experiment. In both settings, our method defends against \textbf{CCE} and maintains model utility. With \textbf{SE}, we achieve better FID scores, as summing \( L_{sub} \) preserves a broader range of knowledge from \( L_{reg} \).

\begin{table}[t]
    \centering
    \small
    \begin{tabular}{lccc}
        \toprule
        \textbf{\textit{Entity}} & \textbf{FID} $\downarrow$ & \textbf{CCE} $\downarrow$ & \textbf{UD} $\downarrow$ \\
        \hline \\[-7pt]
        English Springer & 20.12 / 18.14 & 0.12 / 0.47 & 0.07 / 0.34 \\
        Van Gogh & 20.16/ 17.54 & 0.07 / 0.52 & 0.04 / 0.46\\
        Elon Musk & 20.15 / 18.22 & 0.07 / 0.49 & 0.05 / 0.57\\
        Grumpy Cat & 24.57 / 20.32 & 0.18 / 0.54 & 0.14 / 0.46 \\
        \bottomrule
    \end{tabular}
    \vspace{-2mm}    
    \caption{\textbf{Ablation Study for the base method (ESD / UCE).}}
    \vspace{-3mm}
    \label{tab:BaseMethod}
\end{table}

\subsection{Ablation Studies} \label{sec: Ablation Study}

\minisection{Effect of selecting the \textit{Reference Subspace}.} We present results for four ways of constructing the \textit{Reference Subspace}, including TI tokens from Steps 30, 50, 70 of TI training, and the concept's general term, as shown in Table \ref{tab: Ablation Reference Subspace}. Visualizations of each token are in Fig. \ref{fig:Ablation1}. At Step 30, the token is far from the target, with little relevant information. At Step 50, the token shares features with the target but is not an exact match, enabling minimal transformation. At Step 70, the token is close to the target, making full erasure difficult. These observations align with Table \ref{tab: Ablation Reference Subspace}, where Step 50 balances ASR, FID, and CLIP scores, while Step 30 improves ASR but slightly reduces image quality. Step 70 shows results similar to CA, indicating SuMa’s inefficacy when the \textit{Reference Subspace} and \textit{Target Subspace} are too close.

% \minisection{Effect of choosing the base method}. As mentioned in Section \ref{sec:suma}, we choose CA loss to integrate into our method. Here, we provide the results of using ESD \cite{gandikota2023erasing} and UCE \cite{gandikota2024unified} as the base methods and report the results in Table \ref{tab:BaseMethod}. ESD is very similar to CA, except that ESD is a pushing approach, while CA is a mapping approach. UCE modifies the cross-attention map to erase the concept. In our experiments, we found that modifying the attention layer alone leads to the phenomenon where the token learned by CCE can only generate the target concept when conditioned on the erased model. We observe that ESD works fine in terms of defending against attacks, but the FID moderately increases, while UCE fails to defend against the attack methods. Our conclusion is that we should choose a base method that directly modifies the noise prediction of the model, and that the mapping approach is preferred over the pushing approach. 

\minisection{Effect of choosing the base method.} As noted in Section~\ref{sec:suma}, we adopt CA loss for our method. Results using ESD~\cite{gandikota2023erasing} and UCE~\cite{gandikota2024unified} as base methods are shown in Table~\ref{tab:BaseMethod}. While ESD resembles CA, it uses a pushing strategy, whereas CA employs a mapping strategy. UCE erases concepts by altering the cross-attention map. Our experiments show that modifying only the attention layer causes the token learned by CCE to express the target concept only under the erased model as shown in Appendix \ref{sec:ModifyCrossAttn}. ESD provides some defense but increases FID moderately, while UCE fails to defend. We conclude that modifying the model’s noise prediction is more effective, as it redirects the target concept's trajectory toward another in the original model. The mapping approach is also favored over the pushing one, as discussed in Section~\ref{sec:refspace} and supported by our results.

\minisection{Effect of adding Subspace Regularization.} We present the effect of adding Subspace Regularization and report the results in Table \ref{tab:Ablation Reg Subspace}. We observe that without regularization, our method performs slightly better against the CCE attack, with an average 2\% reduction in attack rate, but at the cost of a gradual increase in the FID score. This phenomenon was studied in \cite{bui2024erasing}, where it was shown that the concept closest to the target is the most affected. Therefore, we conclude that adding Subspace Regularization loss is necessary to maintain the model’s effectiveness.

\section{Conclusion and Limitations}
In this paper, we introduced SuMa, a method that robustly and effectively erases narrow concepts in Text-to-Image Diffusion models. SuMa constructs a \textit{Target Subspace} for the concept to be erased and maps it to a \textit{Reference Subspace}. Experimental results show our method is the first to meet these dual objectives for narrow concepts. A key limitation is its time cost, as it may require many iterations to construct both subspaces. Future work will aim to accelerate this process and improve target concept representation.

%% file: sec/X_sup.tex
\clearpage
\setcounter{page}{1}
\maketitlesupplementary

% In this supplementary material, we first present both quantitative and qualitative results for all the aforementioned concepts when applying $L_{sub}$ alone, along with an intuitive explanation of why applying $L_{sub}$ alone cannot completely erase the concept, in Section \ref{sec:Additional Ablations}. Next, we include an additional experiment on artistic style erasure in Section \ref{Additional Artistic}. We then provide quantitative results for the standard test of verifying the concept-erased models using the target's textual prompts in Section \ref{sec: ASR Target Textual}. Following that, we present the exact quantitative results for each concept listed in Section \ref{sec: Exp Setting}, as detailed in Section \ref{sec:QuantiativeDetail}. This is followed by additional visualizations of Textual Inversion (TI) at different levels of the application method $\mathcal{M}$ and at various training steps, which help illustrate why selecting the \textit{Reference Subspace} at early training steps is effective, as described in Section \ref{sec: More Visualization TI}. Finally, we provide qualitative results for each concept in Section \ref{sec:Additional Qualiatative}.
\input{sec/Appendix/A_Related_work}

\input{sec/Appendix/B_Toy_Experiment}
\input{sec/Appendix/C_Subspace_Pushing}
\input{sec/Appendix/D_Modify_CrossAttn_Only}
\input{sec/Appendix/E_NSFW_Concepts}
\input{sec/Appendix/F_Standard_Test}

\input{sec/Appendix/G_Training_Time}
\input{sec/Appendix/H_TI_Visualization}
\input{sec/Appendix/I_More_Quantiative_Result}
\input{sec/Appendix/J_More_Qualitative_Result}

%% file: sec/Appendix/A_Related_work.tex
\section{Related Work Discussion} \label{sec:Related_Work_Discussion}

\begin{figure}[t]
  \centering
   \includegraphics[width=1.0\linewidth]{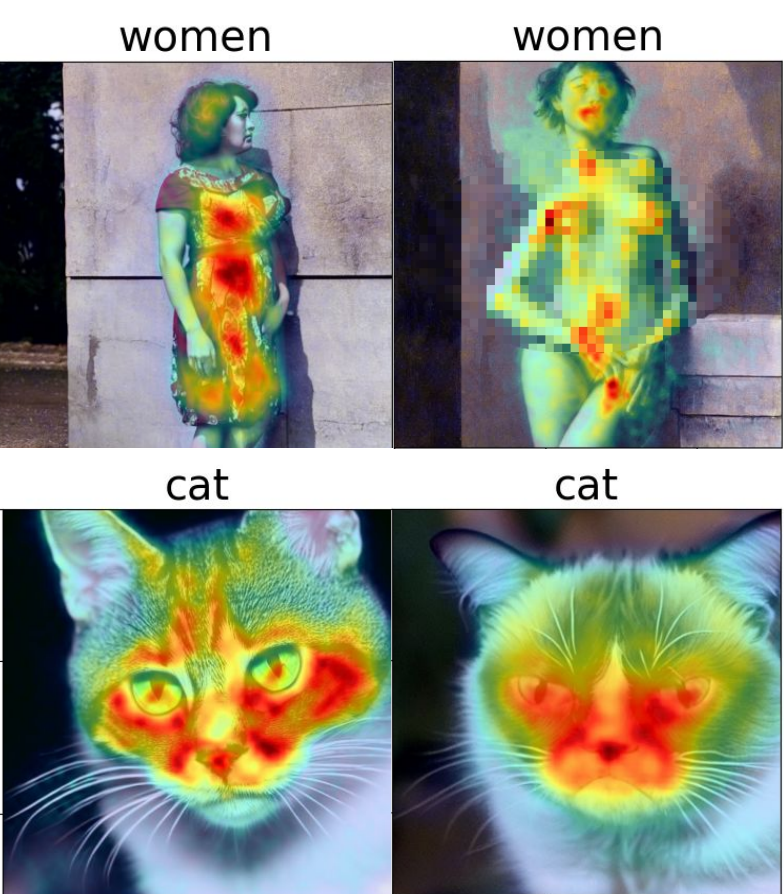}
   \caption{Cross Attention Map Visualization of NSFW concepts and narrow concepts.} 
   \label{fig:Observation_NSFW}
\end{figure}

\begin{figure}[t]
  \centering
   \includegraphics[width=1.0\linewidth]{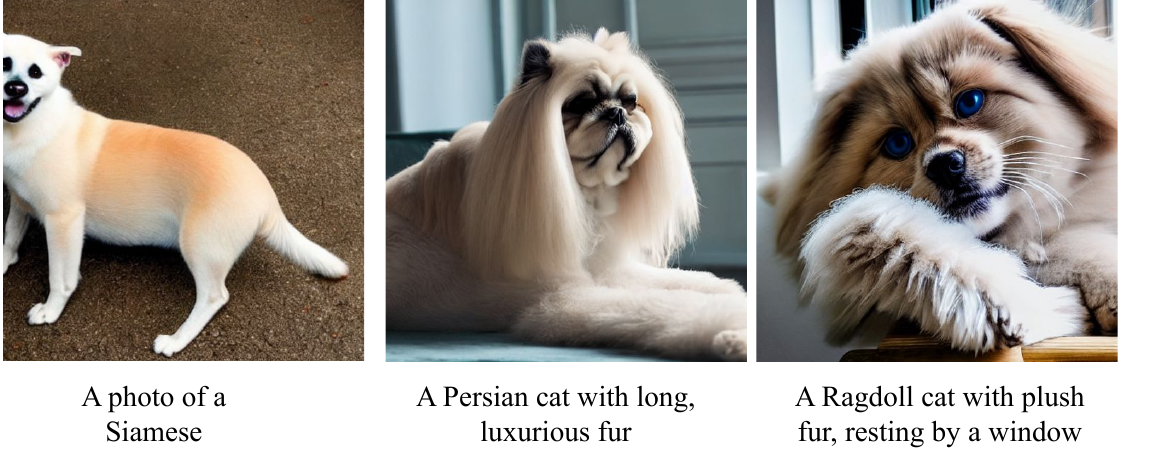}
   \caption{Output of DUO method after erase Grumpy Cat.} 
   \label{fig:DUO_Output}
\end{figure}

\minisection{Circumventing Methods. }Although CCE and UnlearnDiff (UD) optimize the following objective function:
\begin{equation}
    c = \underset{c}{\arg\min} \; \mathbb{E}_{z_t, \epsilon, t} \left[ \| \epsilon - \Phi\big(z_t, \mathcal{T}(c), t\big) \|_2^2 \right],
\end{equation}
the way they find the final token is different. CCE directly uses the latent token found by optimizing this formula as the final token, whereas UD requires a more complicated process because they want to construct a readable adversarial prompt. They first start with \( \mathbf{X} \in \mathbb{R}^{k \times L} \), where \( k \) is the number of tokens we expect to construct the prompt, \( L \) is the length of the vocabulary, and this matrix is a right stochastic matrix. Next, they take the argmax over the columns of \( \mathbf{X} \) and construct a one-hot matrix \( \mathbf{Y} \in \mathbb{R}^{k \times L} \). These operations are made differentiable. Then, they construct the prompt \( c = \mathbf{Y} \times \mathbf{Z} \), where \( \mathbf{Z} \in \mathbb{R}^{L \times d} \) is the token embedding weight matrix, and \( d \) is the token embedding dimension. During the optimization process, they update \( \mathbf{X} \) using Projected Gradient Descent to ensure that \( \mathbf{X} \) remains a right stochastic matrix. We can clearly see that this process is much more complicated than CCE and the search space is narrower than CCE because they need to ensure that \( \mathbf{X} \) remains a right stochastic matrix. This is the main reason why many works have successfully protected the model against UD, but only a few, such as STEREO, DUO, and ours, have successfully protected the model against CCE.

\minisection{Direct Unlearning Optimization (DUO).} \textbf{DUO} is a method proposed to achieve the same objective as ours, but specifically for \textbf{NSFW} concepts. They found that most existing methods focus on the textual space to erase concepts, while attack methods use the image space to create adversarial prompts. Therefore, they propose a method that uses the image space as the foundation. The core technique that makes this method work is the editing approach, where they use this technique to create a pair of images: the first is an inexact image, and the second is the approximate one. They then apply \textbf{Diffusion-DPO} with the goal of assigning a high score to the approximate image and a low score to the inexact one. We found that this method is suitable for \textbf{NSFW} concepts because it is reasonable to edit the ``naked" concept to a ``dressed in" concept. We visualize the attention map of the word ``women" in two prompts: ``A naked woman" and ``A dressed woman" in Figure \ref{fig:Observation_NSFW}. We can see that the attention reflects a clear difference between the two concepts. However, for the narrow concept ``Grumpy Cat", it is reasonable to edit this concept to ``Cat", but as shown in Figure \ref{fig:Observation_NSFW}, the difference between these two concepts is not clear, so the \textbf{DUO} method doesn't work for this kind of concept. We could ask the question: ``Why don't we edit `Grumpy Cat' to `Grumpy Dog'?" We conducted an additional experiment to answer this question and found that after doing so, other kinds of Cat also became a Dog, as shown in Figure \ref{fig:DUO_Output}. Therefore, we conclude that directly applying \textbf{DUO} to narrow concepts is insufficient, and we need a finer-grained method, such as the one we propose, to address the problem of narrow concepts.

%% file: sec/Appendix/B_Toy_Experiment.tex
\section{More Observation}\label{sec:More_Toy_Exp}
We further verify the effect of token similarity on erasing capability. After running the \textbf{Subspace Construction} phase, we end up with a set of target tokens \( \mathcal{S}_1 = \{x_1, x_2, x_3\} \). Then, we use the \textbf{CA} method to randomly erase one token from this set and find that the remaining tokens cannot be erased. However, when we find a new token \( e \) satisfying this equation:

\begin{equation}
e = \underset{e}{\arg\min} \sum_{i=1}^{M} \sum_{j=1}^{3} \left( \| W_k^i e - W_k^i x_j \|_2^2 + \| W_v^i e - W_v^i x_j \|_2^2 \right)
\end{equation}
where \( M \) is the number of U-Net layers, and \( W_k \) and \( W_v \) are the to-\( k \) and to-\( v \) weight matrices in the cross-attention layer, respectively. Specifically, this equation finds a token closest to all three tokens. Using \textbf{CA} to erase \( e \), we found that all tokens in \( \mathcal{S}_1 \) are erased, and the cosine similarity between \( e \) and each token in \( \mathcal{S}_1 \) exceeds 0.6. We noted that erasing this token alone does not help protect the model against \textbf{CCE} or \textbf{UD} attacks. However, it further verifies that the similarity between the erased token and others has a significant effect on the erasure capability.

%% file: sec/Appendix/C_Subspace_Pushing.tex
\section{Subspace Pushing} \label{sec:SubPush}
\begin{table}[t]
    \centering
    \small
    \begin{tabular}{lccc}
        \toprule
        \textbf{\textit{Entity}} & \textbf{FID} $\downarrow$ & \textbf{CCE} $\downarrow$ & \textbf{UD} $\downarrow$ \\
        \hline \\[-7pt]
        English Springer & 21.96 & 0.08 & 0.06  \\
        Van Gogh & 20.27 & 0.03 & 0.04 \\
        Elon Musk & 21.32 & 0.06 & 0.15 \\
        Grumpy Cat & 25.37 & 0.13 & 0.09 \\
        \bottomrule
    \end{tabular}
    \caption{\textbf{Result of using Subspace Pushing approach.}}
    \vspace{-1em}
    \label{tab:Subspace Pushing}
\end{table}

In Section \ref{sec:refspace}, we mentioned that a mapping approach is usually better than a pushing approach, with examples being \textbf{ESD} and \textbf{CA} \cite{lu2024mace, srivatsan2025stereo}, and our method is also a mapping approach. Here, we perform additional experiments with the \textbf{Subspace Pushing} approach. The framework for this experiment is almost identical to \textbf{SuMa}, except that we no longer need the reference subspace. Instead, we try to fine-tune the model such that the target subspace is pushed away from the original one. Specifically, we replace \( L_{proj} \) with the following \( L_{push} \):

\vspace{-1.0em}
\begin{align}
    L_{push} &= \max \left( \tau - \sum_{i=1}^M \sum_{j=1}^l \left\| W_k^i \mathcal{T} \left( p + \langle u \rangle_j \right) - \right. \right. \notag \\
    &\quad \left. \left. \mathbf{P}_{\mathbf{U}^i} W_k^i \mathcal{T} \left( p + \langle u \rangle_j \right) \right\|_2^2, 0 \right) \label{loss: pushing}
\end{align}
where \( \mathbf{P}_{\mathbf{U}^i} \) is the projection matrix onto \( \mathbf{U}^i \), computed similarly to Formula \ref{eq:Proj_matrix}, and \( \tau \) controls how far we want to push the original subspace away and we report the result in Table \ref{tab:Subspace Pushing}. Overall, we can see that compared to Subspace Mapping, Subspace Pushing performs slightly better in terms of protecting the model against adversarial attacks, but at the cost of significantly reducing image quality.

%% file: sec/Appendix/D_Modify_CrossAttn_Only.tex
\section{Modify Cross-Attn Only}\label{sec:ModifyCrossAttn}

\begin{figure}[t]
  \centering
   \includegraphics[width=1.0\linewidth]{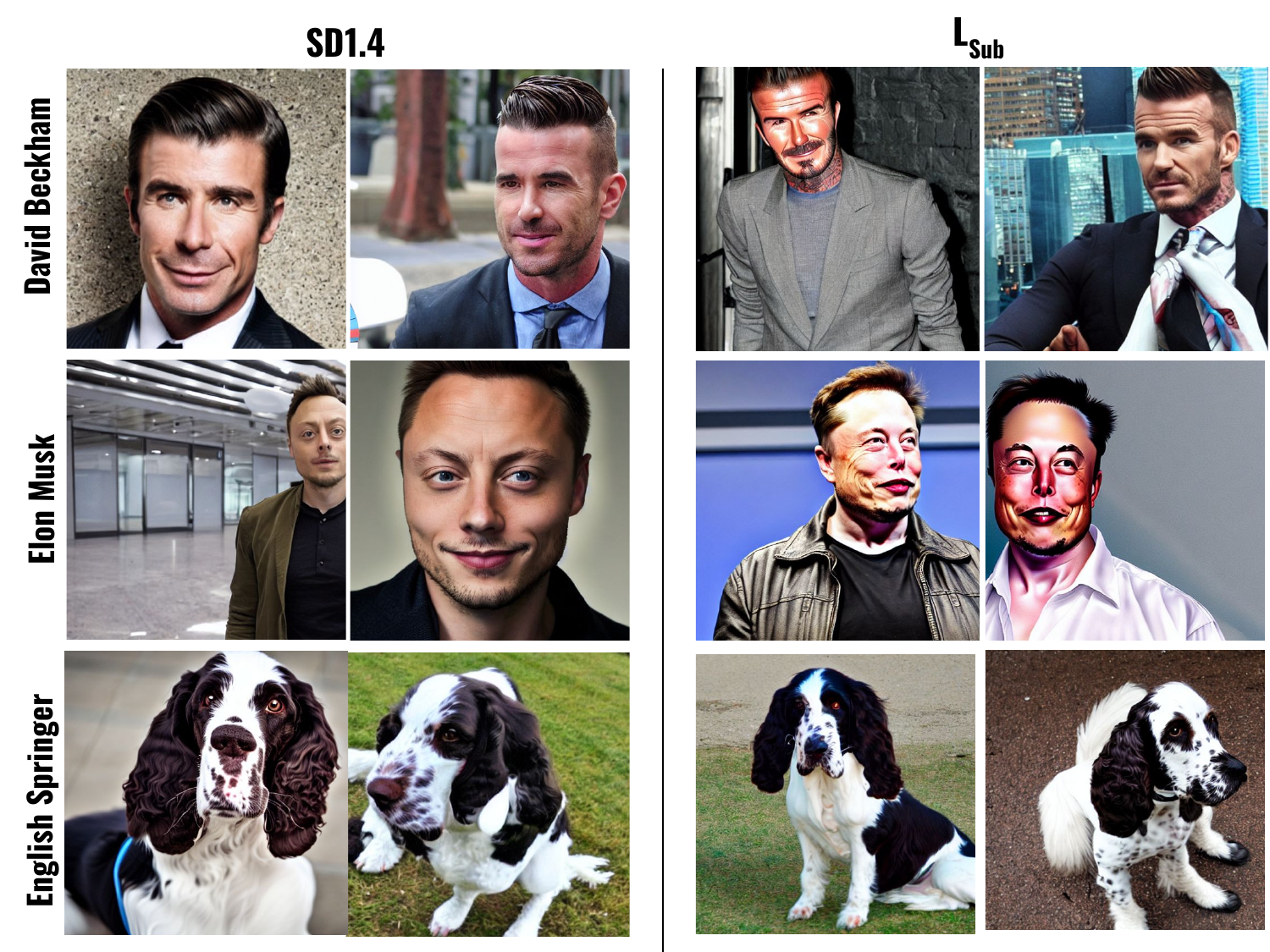}
   \caption{Output of the TI token when passed through SD1.4 and the new model fine-tuned with $L_{sub}$ only.} 
   \label{fig:Additional Ablation Study}
\end{figure}

\begin{table}[t]
  \centering
  \setlength{\tabcolsep}{12pt}
  \begin{tabular}{@{}lcc@{}}
    \toprule
    Concept & ASR ($L_{sub} + L_{CA}$) & ASR ($L_{sub}$) \\ 
    \midrule
    \textit{English Springer} & 0.12 & 0.37\\ 
    \textit{Garbage Truck} & 0.03 & 0.29 \\
    \textit{Golf Ball} & 0.18 & 0.34 \\
    \textit{\textbf{Elon Musk}} & 0.09 & \textbf{0.73} \\
    \textit{\textbf{David Beckham}} & 0.12 & \textbf{0.69} \\
    \textit{\textbf{Adam Lambert}} & 0.05 & \textbf{0.57} \\
    \textit{\textbf{Van Gogh}} & 0.05 & \textbf{0.76} \\
    \textit{Mickey Mouse} & 0.19 & 0.43 \\
    \textit{R2D2 Robot} & 0.14 & 0.47 \\
    \textit{Grumpy Cat} & 0.16 & 0.38 \\
    \bottomrule
  \end{tabular}
  \caption{Comparison of ASRs when applying $L_{sub}$ only versus $L_{sub} + L_{CA}$. The experiments with Identity and Artistic Style Erasure, in which the ASRs when using $L_{sub}$ only are larger than 50\%, are in \textbf{bold}.}
  \label{tab:Additional Ablation Study}
\end{table}

We provide an ablation study by applying $L_{sub}$ alone and in combination with $L_{CA}$, reporting the Attack Success Rate (ASR) in Table \ref{tab:Additional Ablation Study}. In general, compared to existing E-CEM methods, applying $L_{sub}$ alone achieves lower ASRs. In most concepts, the ASR is smaller than 50\%, except for the \textbf{Identity} and \textbf{Artistic Style} categories, where it performs significantly worse than normal. Furthermore, as shown in Figure \ref{fig:Additional Ablation Study}, the TI token for the \textbf{Identity} category from model $\theta^{\prime}$ (fine-tuned using $L_{sub}$ alone) generates the target concept only when conditioned on $\theta^{\prime}$ and fails to generate the target concept when conditioned on $\theta$ (the original model). Interestingly, this phenomenon is observed exclusively in the \textbf{Identity} category and not in others. In our opinion, this happens because Textual Inversion (TI) acts like a process that collects all the residual knowledge of the model about concepts closely related to the target concept and reconstructs the target concept. For the \textbf{Identity} category, TI tokens are more sensitive and constrained on $\theta^{\prime}$, leading to this phenomenon. When combining $L_{sub}$ with $L_{CA}$, $L_{sub}$ acts merely as a tool to protect the fine-tuned model from further CCE attacks. As mentioned in \cite{srivatsan2025stereo}, applying the erasing method $\mathcal{M}$ iteratively can completely erase the target concept, though this results in a significant decrease in image quality. We hypothesize that, intuitively, applying method $\mathcal{M}$ iteratively will help us identify all concepts that can be used to reconstruct the target concept during the process of finding TI tokens. When combined with $L_{sub}$, this method will guide the model to converge to a point where it can completely erase the target concept, with minimal weight modification. Figure \ref{fig:Iterative TI output} illustrates the output of TI at different levels of applying method $\mathcal{M}$ to erase the TI of the target concept.

%% file: sec/Appendix/E_NSFW_Concepts.tex
\section{NSFW Concept Erasure}\label{sec:NSFW_Concepts}
\begin{figure*}[t]
  \centering
   \includegraphics[width=1.0\linewidth]{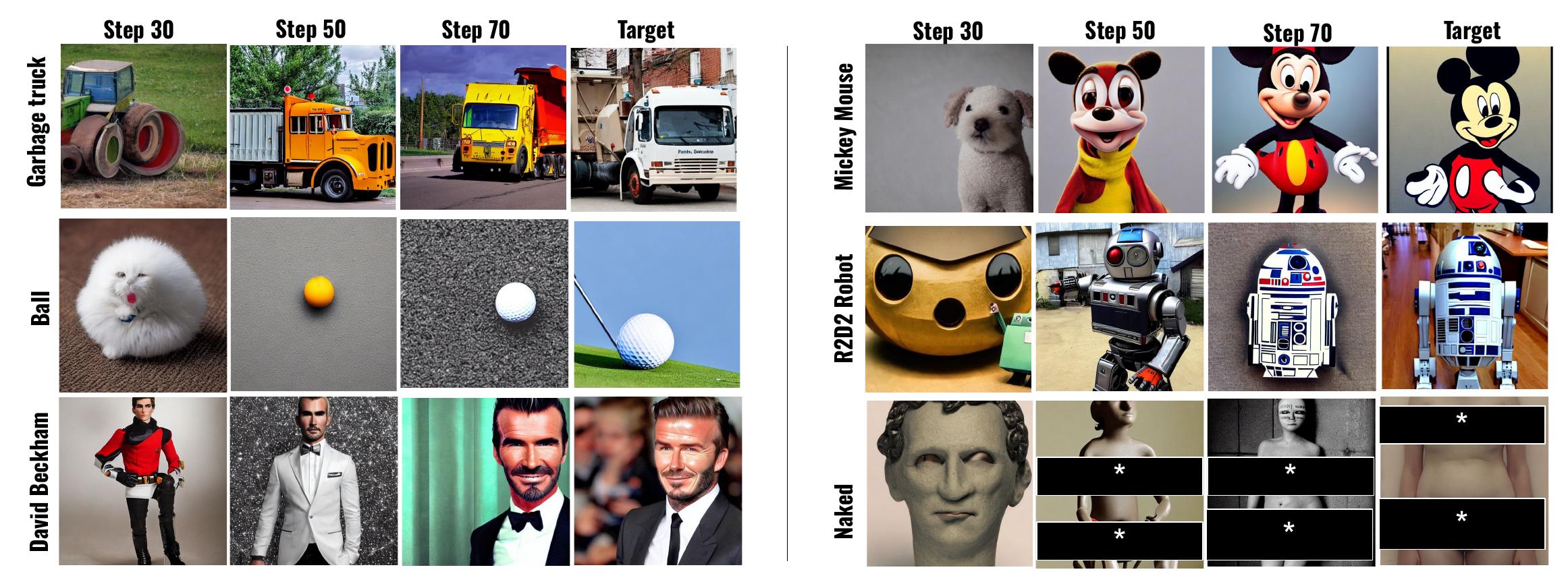}
   \caption{Output of the TI token at different TI training steps.} 
   \label{fig:Step TI output}
\end{figure*}

\begin{table}[t]
    \centering
    \small
    \resizebox{0.47\textwidth}{!}{%
    \begin{tabular}{lccc}
        \toprule
        \textbf{\textit{Mehod}} & \textbf{FID} $\downarrow$ & \textbf{CCE} $\downarrow$ & \textbf{UD} $\downarrow$ \\
        \hline \\[-7pt]
        Subspace Pushing & 22.12 & 0.12 & 0.09  \\
        DUO & 17.56 & 0.07 & 0.06 \\
        DUO + SuMa (English Springer) & 18.21 & 0.05 & 0.04 \\
        DUO + SuMa (Elon Musk) & 18.94 & 0.04 & 0.03 \\
        DUO + SuMa (Grumpy Cat) & 23.24 & 0.03 & 0.02 \\

        \bottomrule
    \end{tabular}
    } % End of resizebox
    \caption{\textbf{Result of erasing ``Nudity'' concept from SD1.4.}}
    \vspace{-1em}
    \label{tab:NSFW}
\end{table}

As mentioned in Section \ref{sec: Exp Setting}, SuMa doesn't work very well for NSFW concepts, such as the "Nudity" concept. The reason is that during Textual Inversion fine-tuning, the early steps for the ``Nudity" concept directly converge to this concept, as shown in Figure \ref{fig:Step TI output}, making it impossible to construct a reference subspace for this concept using our proposed approach. Instead, we propose two alternative solutions. First, we could employ the Pushing Approach, as described in Section \ref{sec:SubPush}, because this approach does not require a reference subspace. However, in our experiments, we found that while this approach is able to keep the model safe under CCE and UD attacks (Following previous work, we use the Nudenet Detector to compute ASR      
), its FID and CLIP scores significantly decrease, as shown in Table \ref{tab:NSFW}. Second, we found that after erasing one narrow concept based on our method, we can still apply DUO afterward, with results very close to applying DUO directly, as shown in Table \ref{tab:NSFW}. So, in conclusion, our work could be combined with DUO to eliminate all kinds of concepts and advance the development of a safe text-to-image model.

%% file: sec/Appendix/F_Standard_Test.tex
\section{Standard Test on the Target's Textual Prompts} \label{sec:StandartTest}
\begin{table*}[t]
    \centering
    \resizebox{\textwidth}{!}{
    \setlength{\tabcolsep}{5pt}
    \small
    \begin{tabular}{lcccccccccccc}
        \toprule
        \textbf{Method} & \multicolumn{1}{c}{\textbf{Artistic}} & \multicolumn{3}{c}{\textbf{Subclass}} & \multicolumn{3}{c}{\textbf{Identity}} & \multicolumn{3}{c}{\textbf{Instance}} \\
        \cmidrule(lr){2-2} \cmidrule(lr){3-5} \cmidrule(lr){6-8} \cmidrule(lr){9-11}
        & VanGogh & English Springer & Garbage Truck & Golf Ball & \textit{David Beckham} & \textit{Elon Musk} & \textit{Adam Lambert} & Grumpy Cat & R2D2 & Mickey \\
        \midrule
        SD \cite{RobinSD1.4} & 0.86 / 0.85 & 0.86 / 0.87 & 0.74 / 0.71 & 0.90 / 0.85 & 0.84 / 0.81 & 0.91 / 0.89 & 0.86 / 0.83 & 0.97 / 0.94 & 0.96 / 0.95 & 0.96 / 0.94 \\
        \midrule
        CA \cite{kumari2023ablating} & 0.06 / 0.03 & 0.00 / 0.01 & 0.03 / 0.02 & 0.18 / 0.16 & 0.04 / 0.03 & 0.03 / 0.02 & 0.05 / 0.04 & 0.00 / 0.00 & 0.00 / 0.00 & 0.00 / 0.00 \\
        MACE \cite{lu2024mace} & 0.02 / 0.01 & 0.00 / 0.01 & 0.07 / 0.05 & 0.02 / 0.03 & 0.03 / 0.02 & 0.00 / 0.01 & 0.01 / 0.00 & 0.00 / 0.00 & 0.00 / 0.00 & 0.00 / 0.00 \\
        RACE \cite{kim2024race} & 0.07 / 0.06 & 0.00 / 0.01 & 0.09 / 0.07 & 0.00 / 0.00 & 0.02 / 0.01 & 0.06 / 0.05 & 0.03 / 0.02 & 0.00 / 0.00 & 0.00 / 0.00 & 0.00 / 0.00 \\
        AdvUnlearn \cite{zhang2024defensive} & 0.04 / 0.02 & 0.00 / 0.00 & 0.03 / 0.01 & 0.02 / 0.00 & 0.05 / 0.03 & 0.02 / 0.01 & 0.00 / 0.00 & 0.00 / 0.00 & 0.12 / 0.09 & 0.00 / 0.01 \\
        DUO \cite{parkdirect} & 0.02 / 0.01 & 0.03 / 0.02 & 0.02 / 0.04 & 0.03 / 0.05 & 0.02 / 0.02 & 0.03 / 0.02 & 0.01 / 0.01 & 0.04 / 0.03 & 0.08 / 0.06 & 0.05 / 0.02 \\
        STEREO \cite{srivatsan2025stereo} & 0.00 / 0.00 & 0.00 / 0.00 & 0.02 / 0.01 & 0.01 / 0.00 & 0.00 / 0.00 & 0.00 / 0.00 & 0.01 / 0.00 & 0.00 / 0.00 & 0.00 / 0.00 & 0.00 / 0.00 \\
        Ours & 0.04 / 0.03 & 0.00 / 0.00 & 0.01 / 0.02 & 0.03 / 0.01 & 0.01 / 0.01 & 0.03 / 0.02 & 0.01 / 0.01 & 0.00 / 0.00 & 0.00 / 0.00 & 0.00 / 0.00 \\
        \bottomrule
    \end{tabular}
    }
    \caption{\textbf{ASR of Artistic Style and Target's Textual (SD1.4 / SD1.5)}}
    \label{tab: Combined ASR}
\end{table*}

\begin{table}[t]
    \centering
    \resizebox{0.5\textwidth}{!}{
    \setlength{\tabcolsep}{10pt} 
    \small
    \begin{tabular}{lccccc} % Changed to 'lccccc' for left-aligned first column
        \toprule
        \textbf{Metric} & \textbf{SuMa} & \textbf{STEREO} & \textbf{RACE} & \textbf{AdvUnlearn} & \textbf{AC} \\
        \midrule
        TT (Minute) & 23 & 23 & 15 & 15 & 12 \\
        MU (GB)     & 12.21 & 12.74 & 11.67 & 12.51 & 11.19 \\
        \bottomrule
    \end{tabular}
    }
    \caption{\textbf{Training Time (TT) and Memory Usage (MU) for Different Methods}}
    \label{tab: TT MU}
\end{table}

In this section, we conduct the standard test to see if the concept-erased models could avoid generating the target concept when using prompts with the target terms.
Following \cite{kumari2023ablating}, we use ChatGPT to generate 100 prompts for each concept and apply the same classifier method mentioned in Section \ref{sec: Exp Setting} to evaluate the ASR or the unsuccessful erasure rate in this case. We report the results in Table \ref{tab: Combined ASR}. In general, this is a trivial task, and all methods perform very well. For CA, there was a minor issue with the concept of \textit{Golf Ball} and AdvUnlearn got a problem with \textit{Mickey Mouse}. However, our method, which is based on CA, successfully erased \textit{Golf Ball} and \textit{Mickey Mouse} with the help of $L_{sub}$. Thus, we can claim that our method not only enhances CA by protecting it from CCE attacks but also makes it more robust in trivial erasure tasks.

%% file: sec/Appendix/G_Training_Time.tex
\section{Training Time and Memory Usage.}\label{sec:TrainingTime}
We provide the training time (TT) and memory usage (MU) of different methods in Table \ref{tab: TT MU}. Overall, compared to \textbf{STEREO}, our method consumes the same amount of time and memory. Compared to others, our method takes one and a half times longer but is still within an acceptable range and could scale up.

%% file: sec/Appendix/H_TI_Visualization.tex
\section{More Visualizations of Textual Inversion} \label{sec:TI_Visualization}

\begin{figure*}[t]
  \centering
   \includegraphics[width=1.0\linewidth]{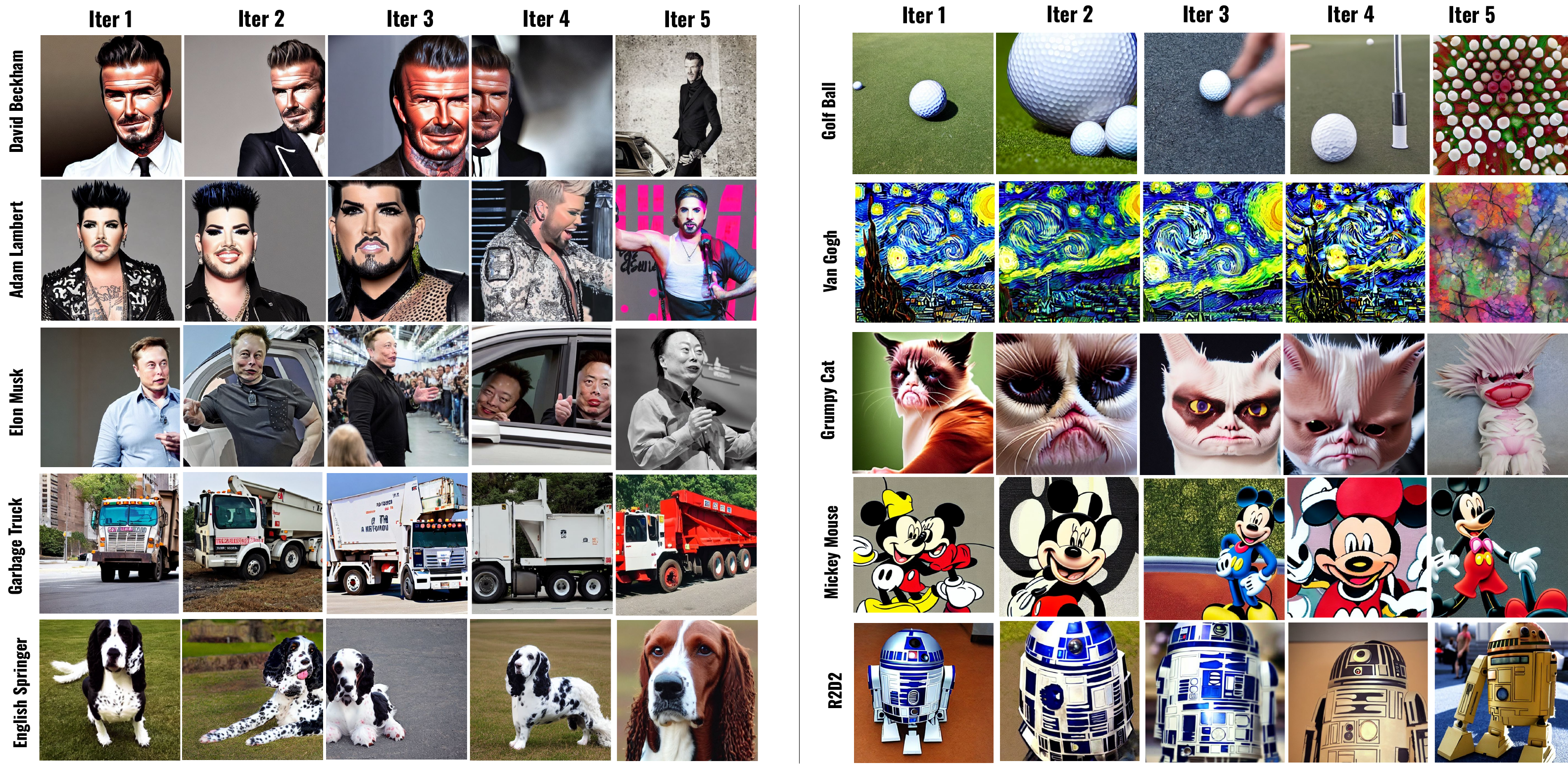}
   \caption{Output of the TI token at different rounds of applying method $\mathcal{M}$} 
   \label{fig:Iterative TI output}
\end{figure*}

We present the output of TI for each concept at different rounds of applying the method $\mathcal{M}$ in Figure \ref{fig:Iterative TI output}. These concepts are all generated by SD1.4. From the first four iterations, the tokens contain significant information across all concepts. By the fifth iteration, the concepts in the \textit{\textbf{Identity}} category are likely removed. Similar observations can be made for \textit{Golf Ball}, \textit{Van Gogh}, and \textit{Grumpy Cat}. In contrast, the other concepts still appear to retain target concept information, but these features are overlapped with those of previous tokens. In our experiment, when we eliminate the subspace constructed by the first three TI tokens, all subsequent tokens at later steps are effectively erased, even if they still contain information about the target concept. We also provide the output of TI at different Textual Inversion training steps in Figure \ref{fig:Step TI output}, as mentioned in Section \ref{sec: Ablation Study}. Early training steps often contain little to no information. The middle steps capture the general meaning of the target concept, including some similar features, but not an exact match. The later steps produce TI tokens much closer to the target concept, as verified in Figure \ref{fig:Step TI output}.

%% file: sec/Appendix/I_More_Quantiative_Result.tex
\section{Quantitative Details} \label{sec:Quantiative_Details}
We present quantitative results for each concept in the \textbf{Subclass}, \textbf{Identity}, and \textbf{Instance} categories in Table \ref{tab: Detail Subclass}, \ref{tab: Detail Identity}, and \ref{tab: Detail Instance}, respectively. For the \textbf{Artistic Style} category, the results are identical to those in Table \ref{tab: Main Result}, as we tested only one artistic style, Van Gogh.

\begin{table*}[t]
    \centering
    \resizebox{\textwidth}{!}{%
    \setlength{\tabcolsep}{12pt}
    \small
    \begin{tabular}{lcccccccccccc}
        \toprule
        \textbf{Method} & \multicolumn{4}{c}{\textbf{English Springer}} & \multicolumn{4}{c}{\textbf{Garbage Truck}} & \multicolumn{4}{c}{\textbf{Golf Ball}} \\
        \cmidrule(lr){2-5} \cmidrule(lr){6-9} \cmidrule(lr){10-13}
        & CCE $\downarrow$ & UD $\downarrow$ & FID $\downarrow$ & CLIP $\uparrow$ & CCE $\downarrow$ & UD $\downarrow$ & FID $\downarrow$ & CLIP $\uparrow$ & CCE $\downarrow$ & UD $\downarrow$ & FID $\downarrow$ & CLIP $\uparrow$ \\
        \midrule
        SD1.4 \cite{RobinSD1.4} & \failed{0.86 / 0.85} & \failed{0.76 / 0.76} & 17.04 / 16.95 & 0.33 / 0.32 & \failed{0.74 / 0.86} & \failed{0.75 / 0.74} & 17.04 / 16.95 & 0.33 / 0.32 & \failed{0.90 / 0.81} & \failed{0.71 / 0.77} & 17.04 / 16.95 & 0.33 / 0.32 \\
        \midrule
        CA \cite{kumari2023ablating} & \failed{0.87 / 0.75} & \failed{0.70 / 0.60} & 19.44 / 19.23 & 0.32 / 0.32 & \failed{0.69 / 0.72} & \failed{0.69 / 0.63} & 19.08 / 19.28 & 0.30 / 0.33 & \failed{0.76 / 0.78} & \failed{0.62 / 0.63} & 19.31 / 19.26 & 0.32 / 0.31 \\
        MACE \cite{lu2024mace} & \failed{0.91 / 0.83} & \failed{0.73 / 0.71} & 16.36 / 16.98 & 0.30 / 0.32 & \failed{0.81 / 0.83} & \failed{0.71 / 0.70} & 16.09 / 16.96 & 0.28 / 0.31 & \failed{0.85 / 0.86} & \failed{0.75 / 0.64} & 16.61 / 16.99 & 0.27 / 0.30 \\
        RACE \cite{kim2024race} & \failed{0.80 / 0.74} & \success{0.20 / 0.19} & 26.57 / 26.13 & 0.28 / 0.28 & \failed{0.68 / 0.73} & \success{0.19 / 0.20} & 26.25 / 26.32 & 0.26 / 0.27 & \failed{0.73 / 0.72} & \success{0.12 / 0.15} & 26.18 / 26.21 & 0.25 / 0.29 \\
        AdvUnlearn \cite{zhang2024defensive} & \failed{0.83 / 0.76} & \success{0.17 / 0.19} & 18.81 / 18.35 & 0.33 / 0.31 & \failed{0.68 / 0.74} & \success{0.18 / 0.15} & 18.08 / 18.35 & 0.30 / 0.30 & \failed{0.76 / 0.72} & \success{0.18 / 0.14} & 18.50 / 18.46 & 0.31 / 0.31 \\
        DUO \cite{parkdirect} & \failed{0.68 / 0.63} & \failed{0.63 / 0.63} & 17.11 / 17.01 & 0.30 / 0.30 & \failed{0.67 / 0.61} & \failed{0.60 / 0.68} & 17.11 / 17.09 & 0.30 / 0.30 & \failed{0.60 / 0.65} & \failed{0.63 / 0.65} & 16.99 / 16.87 & 0.28 / 0.30 \\
        STEREO \cite{srivatsan2025stereo} & \success{0.06 / 0.03} & \success{0.04 / 0.02} & 27.48 / 27.20 & 0.30 / 0.28 & \success{0.02 / 0.04} & \success{0.05 / 0.06} & 27.22 / 27.21& 0.28 / 0.27 & \success{0.04 / 0.02} & \success{0.06 / 0.04} & 27.27 / 27.19 & 0.30 / 0.29 \\
        \rowcolor{blue!10} Ours & \success{0.11 / 0.07} & \success{0.09 / 0.03} & 18.64 / 18.23 & 0.31 / 0.30& \success{0.03 / 0.12} & \success{0.10 / 0.05} & 17.94 / 17.97 & 0.29 / 0.30& \success{0.18 / 0.10} & \success{0.05 / 0.10} & 18.34 / 18.23 & 0.32 / 0.31\\
        \bottomrule
    \end{tabular}%
    } % End of \resizebox
    \caption{\textbf{Evaluation of Erasing across 3 Concepts in the \textbf{Subclass} Category.} With ASRs, we highlight the successful attacks (ASR $\ge$ 20\%) in \failed{red} and the defeated attacks (ASR $<$ 20\%) in \success{green}.}
    \label{tab: Detail Subclass}
\end{table*}

\begin{table*}[t]
    \centering
    \resizebox{\textwidth}{!}{%
    \setlength{\tabcolsep}{12pt}
    \small
    \begin{tabular}{lcccccccccccc}
        \toprule
        \textbf{Method} & \multicolumn{4}{c}{\textbf{David Beckham}} & \multicolumn{4}{c}{\textbf{Elon Musk}} & \multicolumn{4}{c}{\textbf{Adam Lambert}} \\
        \cmidrule(lr){2-5} \cmidrule(lr){6-9} \cmidrule(lr){10-13}
        & CCE $\downarrow$ & UD $\downarrow$ & FID $\downarrow$ & CLIP $\uparrow$ & CCE $\downarrow$ & UD $\downarrow$ & FID $\downarrow$ & CLIP $\uparrow$ & CCE $\downarrow$ & UD $\downarrow$ & FID $\downarrow$ & CLIP $\uparrow$ \\
        \midrule
        SD1.4 \cite{RobinSD1.4} & \failed{0.89 / 0.93} & \failed{0.86 / 0.86} & 17.04 / 16.95 & 0.33 / 0.32 & \failed{0.90 / 0.89} & \failed{0.84 / 0.82} & 17.04 / 16.95 & 0.33 / 0.32 & \failed{0.94 / 0.88} & \failed{0. 88 / 0.84} & 17.04 / 16.95 & 0.33 / 0.32 \\
        \midrule
        CA \cite{kumari2023ablating} & \failed{0.92 / 0.89} & \failed{0.80 / 0.74} & 18.34 / 18.19 & 0.32 / 0.30 & \failed{0.77 / 0.87} & \failed{0.79 / 0.75} & 18.14 / 18.20& 0.30 / 0.30 & \failed{0.90 / 0.91} & \failed{0.77 / 0.79} & 18.13 / 18.42 & 0.30 / 0.31 \\
        MACE \cite{lu2024mace} & \failed{0.89 / 0.89} & \failed{0.62 / 0.65} & 16.84 / 16.61 & 0.30 / 0.28 & \failed{0.87 / 0.85} & \failed{0.63 / 0.66} & 16.58 / 16.68 & 0.28 / 0.28 & \failed{0.89 / 0.87} & \failed{0.70 / 0.59} & 16.79 / 16.78 & 0.27 / 0.27 \\
        RACE \cite{kim2024race} & \failed{0.80 / 0.78} & \failed{0.44 / 0.43} & 24.61 / 24.16 & 0.27 / 0.29 & \failed{0.76 / 0.80} & \failed{0.49 / 0.45} & 24.28 / 24.15 & 0.26 / 0.30 & \failed{0.76 / 0.79} & \failed{0.48 / 0.50} & 24.05 / 24.20 & 0.26 / 0.29 \\
        AdvUnlearn \cite{zhang2024defensive} & \failed{0.91 / 0.69} & \failed{0.53 / 0.51} & 17.55 / 17.81 & 0.31 / 0.30 & \failed{0.66 / 0.67} & \failed{0.54 / 0.56} & 17.39 / 17.62 & 0.29 / 0.31 & \failed{0.72 / 0.71} & \failed{0.49 / 0.53} & 17.63 / 17.64 & 0.29 / 0.31 \\
        DUO \cite{parkdirect} & \failed{0.73 / 0.72} & \failed{0.66 / 0.72} & 17.41 / 17.17 & 0.28 / 0.30 & \failed{0.71 / 0.70} & \failed{0.65 / 0.73} & 17.34 / 17.10& 0.29 / 0.29 & \failed{0.75 / 0.71} & \failed{0.70 / 0.75} & 17.21 / 17.06 & 0.29 / 0.31 \\
        STEREO \cite{srivatsan2025stereo} & \success{0.00 / 0.01} & \success{0.03 / 0.00} & 26.40 / 26.14 & 0.30 / 0.29 & \success{0.04 / 0.00} & \success{0.04 / 0.02} & 26.01 / 26.22 & 0.27 / 0.28 & \success{0.02 / 0.02} & \success{0.02 / 0.01} & 26.56 / 26.09 & 0.29 / 0.27 \\
        \rowcolor{blue!10} Ours & \success{0.12 / 0.08} & \success{0.19 / 0.13} & 18.09 / 17.99 & 0.31 / 0.30 & \success{0.18 / 0.07} & \success{0.19 / 0.15} & 17.79 / 17.88 & 0.29 / 0.29 & \success{0.05 / 0.06} & \success{0.15 / 0.14} & 17.93 / 17.80 & 0.31 / 0.29 \\
        \bottomrule
    \end{tabular}%
    } % End of \resizebox
    \caption{\textbf{Evaluation of Erasing across 3 Concepts in the \textbf{Identity} Category.} With ASRs, we highlight the successful attacks (ASR $\ge$ 20\%) in \failed{red} and the defeated attacks (ASR $<$ 20\%) in \success{green}.}
    \label{tab: Detail Identity}
\end{table*}

\begin{table*}[t]
    \centering
    \resizebox{\textwidth}{!}{%
    \setlength{\tabcolsep}{12pt}
    \small
    \begin{tabular}{lcccccccccccc}
        \toprule
        \textbf{Method} & \multicolumn{4}{c}{\textbf{Mickey Mouse}} & \multicolumn{4}{c}{\textbf{R2D2 Robot}} & \multicolumn{4}{c}{\textbf{Grumpy Cat}} \\
        \cmidrule(lr){2-5} \cmidrule(lr){6-9} \cmidrule(lr){10-13}
        & CCE $\downarrow$ & UD $\downarrow$ & FID $\downarrow$ & CLIP $\uparrow$ & CCE $\downarrow$ & UD $\downarrow$ & FID $\downarrow$ & CLIP $\uparrow$ & CCE $\downarrow$ & UD $\downarrow$ & FID $\downarrow$ & CLIP $\uparrow$ \\
        \midrule
        SD1.4 \cite{RobinSD1.4} & \failed{0.97 / 0.96} & \failed{0.94 / 0.96} & 17.04 / 16.95 & 0.33 / 0.32 & \failed{0.96 / 0.92} & \failed{0.93 / 0.93} & 17.04 / 16.95 & 0.33 / 0.32 & \failed{0.96 / 0.94} & \failed{0.95 / 0.91} & 17.04 / 16.95 & 0.33 / 0.32 \\
        \midrule
        CA \cite{kumari2023ablating} & \failed{0.96 / 0.95} & \failed{0.89 / 0.91} & 18.71 / 18.11 & 0.30 / 0.32 & \failed{0.95 / 0.93} & \failed{0.94 / 0.92} & 18.31 / 18.40 & 0.28 / 0.30 & \failed{0.97 / 0.90} & \failed{0.93 / 0.84} & 18.26 / 18.39 & 0.30 / 0.31 \\
        MACE \cite{lu2024mace} & \failed{0.95 / 0.91} & \failed{0.93 / 0.89} & 17.01 / 16.76 & 0.30 / 0.30 & \failed{0.98 / 0.95} & \failed{0.95 / 0.89} & 16.72 / 16.86 & 0.28 / 0.32 & \failed{0.97 / 0.90} & \failed{0.91 / 0.89} & 16.96 / 16.77 & 0.30 / 0.31 \\
        RACE \cite{kim2024race} & \failed{0.97 / 0.93} & \success{0.07 / 0.05} & 24.43 / 24.25 & 0.28 / 0.30 & \failed{0.93 / 0.90} & \success{0.03 / 0.07} & 24.06 / 24.06 & 0.28 / 0.29 & \failed{0.93 / 0.90} & \success{0.07 / 0.09} & 24.46 / 24.13 & 0.27 / 0.28 \\
        AdvUnlearn \cite{zhang2024defensive} & \failed{0.97 / 0.92} & \success{0.03 / 0.04} & 18.74 / 18.16 & 0.32 / 0.32 & \failed{0.93 / 0.90} & \success{0.03 / 0.03} & 18.25 / 18.18 & 0.31 / 0.33 & \failed{0.93 / 0.97} & \success{0.04 / 0.05} & 18.10 / 18.26 & 0.32 / 0.33 \\
        DUO \cite{parkdirect} & \failed{0.93 / 0.88} & \failed{0.89 / 0.90} & 18.36 / 18.82 & 0.30 / 0.30 & \failed{0.93 / 0.88} & \failed{0.86 / 0.88} & 18.32 / 18.91 & 0.29 / 0.30 & \failed{0.90 / 0.97} & \failed{0.92 / 0.83} & 18.25 / 19.03 & 0.31 / 0.33 \\
        STEREO \cite{srivatsan2025stereo} & \success{0.01 / 0.02} & \success{0.08 / 0.03} & 27.31 / 26.99 & 0.32 / 0.32 & \success{0.05 / 0.02} & \success{0.04 / 0.06} & 26.69 / 26.80 & 0.29 / 0.31 & \success{0.03 / 0.02} & \success{0.03 / 0.01} & 27.10 / 26.91 & 0.31 / 0.33  \\
        \rowcolor{blue!10} Ours & \success{0.19 / 0.15} & \success{0.19 / 0.18} & 22.50 / 22.15 & 0.29 / 0.31 & \success{0.14 / 0.15} & \success{0.18 / 0.16} & 22.19 / 22.17 & 0.29 / 0.29 & \success{0.16 / 0.16} & \success{0.20 / 0.15} & 22.32 22.28 & 0.30 / 0.30 \\
        \bottomrule
    \end{tabular}%
    } % End of \resizebox
    \caption{\textbf{Evaluation of Erasing across 3 Concepts in \textbf{Instance} Category.} With ASRs, we highlight the successful attacks (ASR $\ge$ 20\%) in \failed{red} and the defeated attacks (ASR $<$ 20\%) in \success{green}.}
    \label{tab: Detail Instance}
\end{table*}

%% file: sec/Appendix/J_More_Qualitative_Result.tex
\section{Additional Qualitative Results} \label{sec:Additional Qualiatative}

\begin{figure*}[t]
  \centering
   \includegraphics[width=1.0\linewidth]{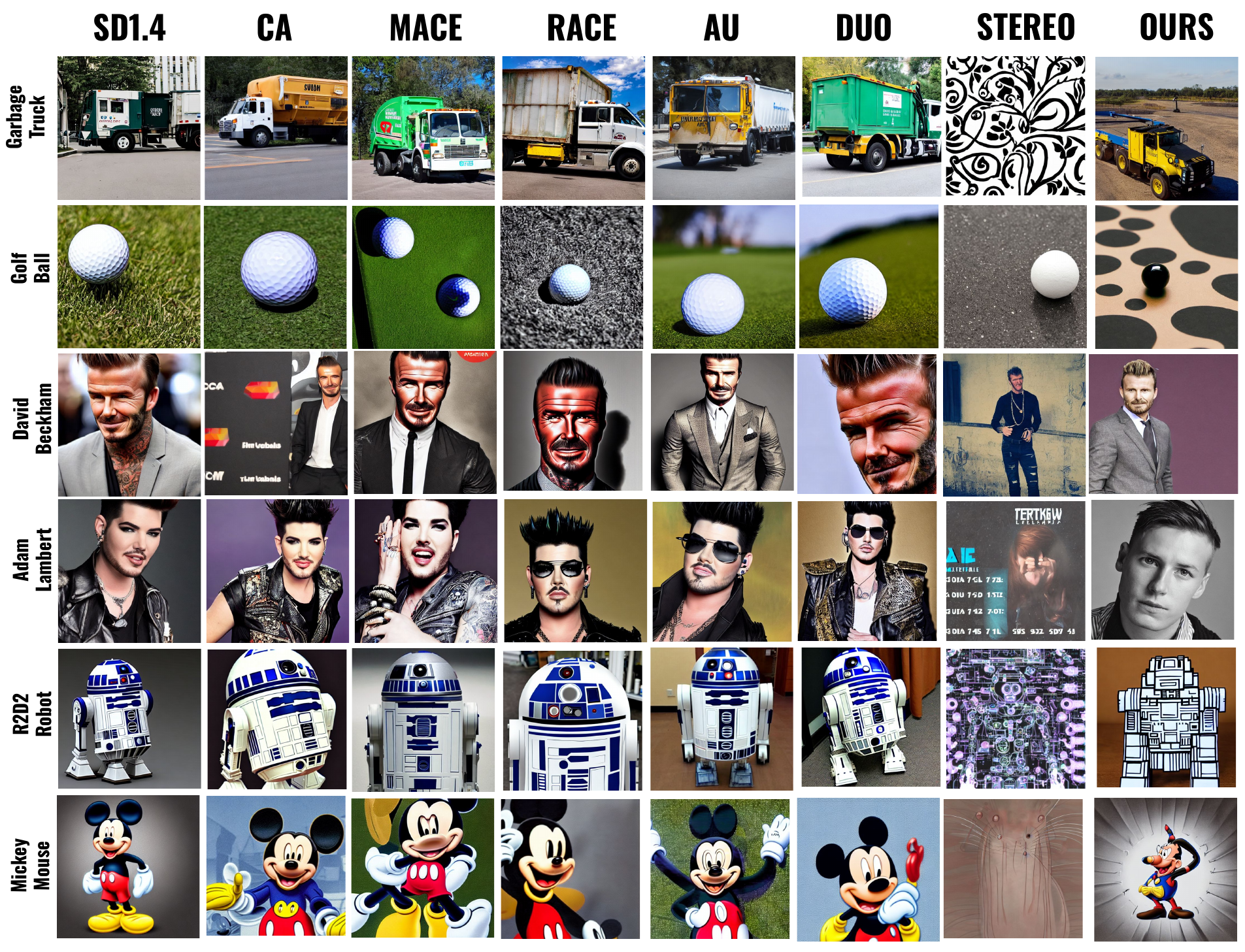}
   \caption{Output of different methods under CCE attack} 
   \label{fig:Remain Attack Output}
\end{figure*}

\begin{figure*}[t]
  \centering
   \includegraphics[width=1.0\linewidth]{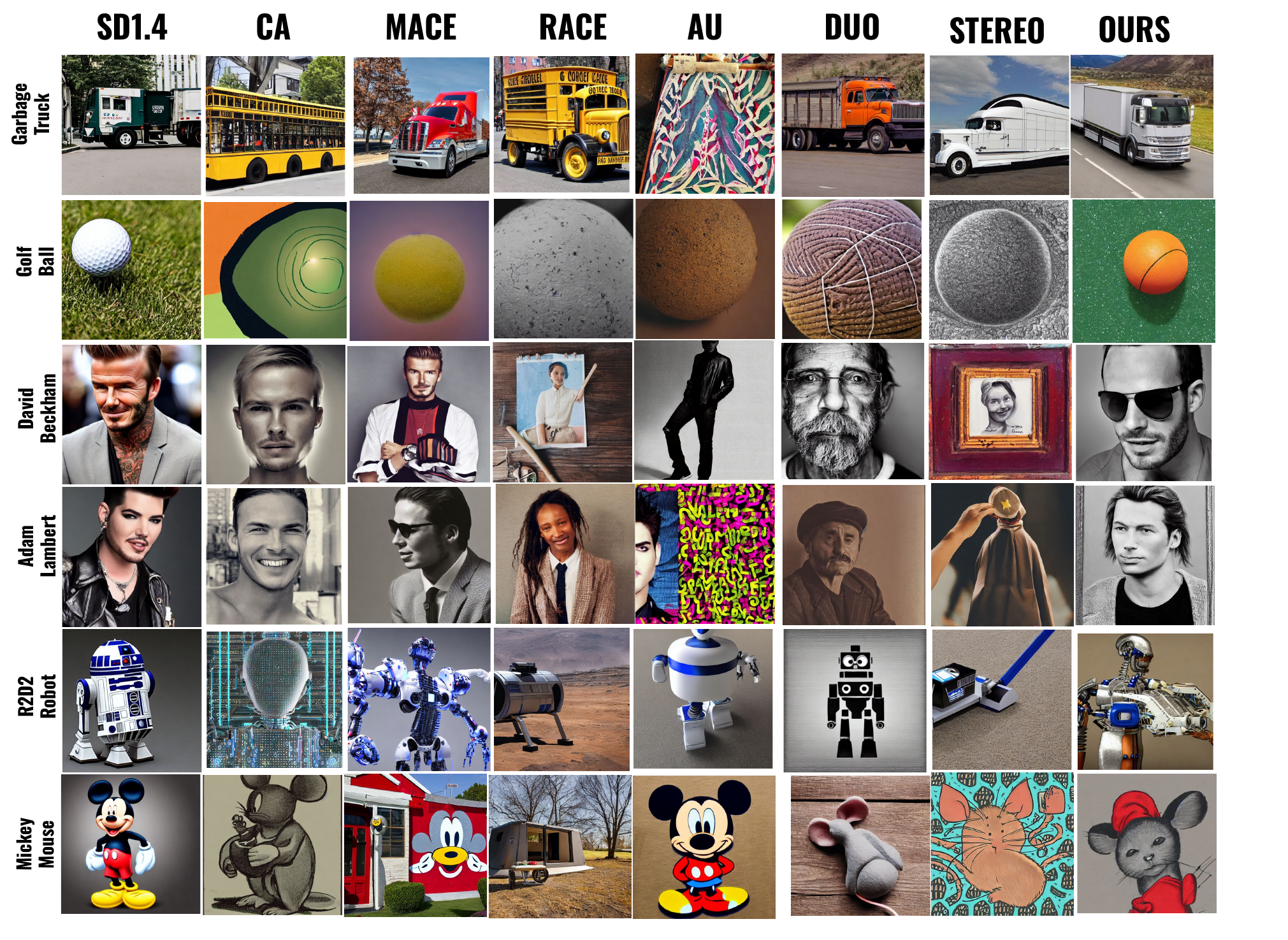}
   \caption{Outputs of different methods for textual prompts} 
   \label{fig:Remain Textual Output}
\end{figure*}

\begin{figure*}[t]
  \centering
   \includegraphics[width=1.0\linewidth]{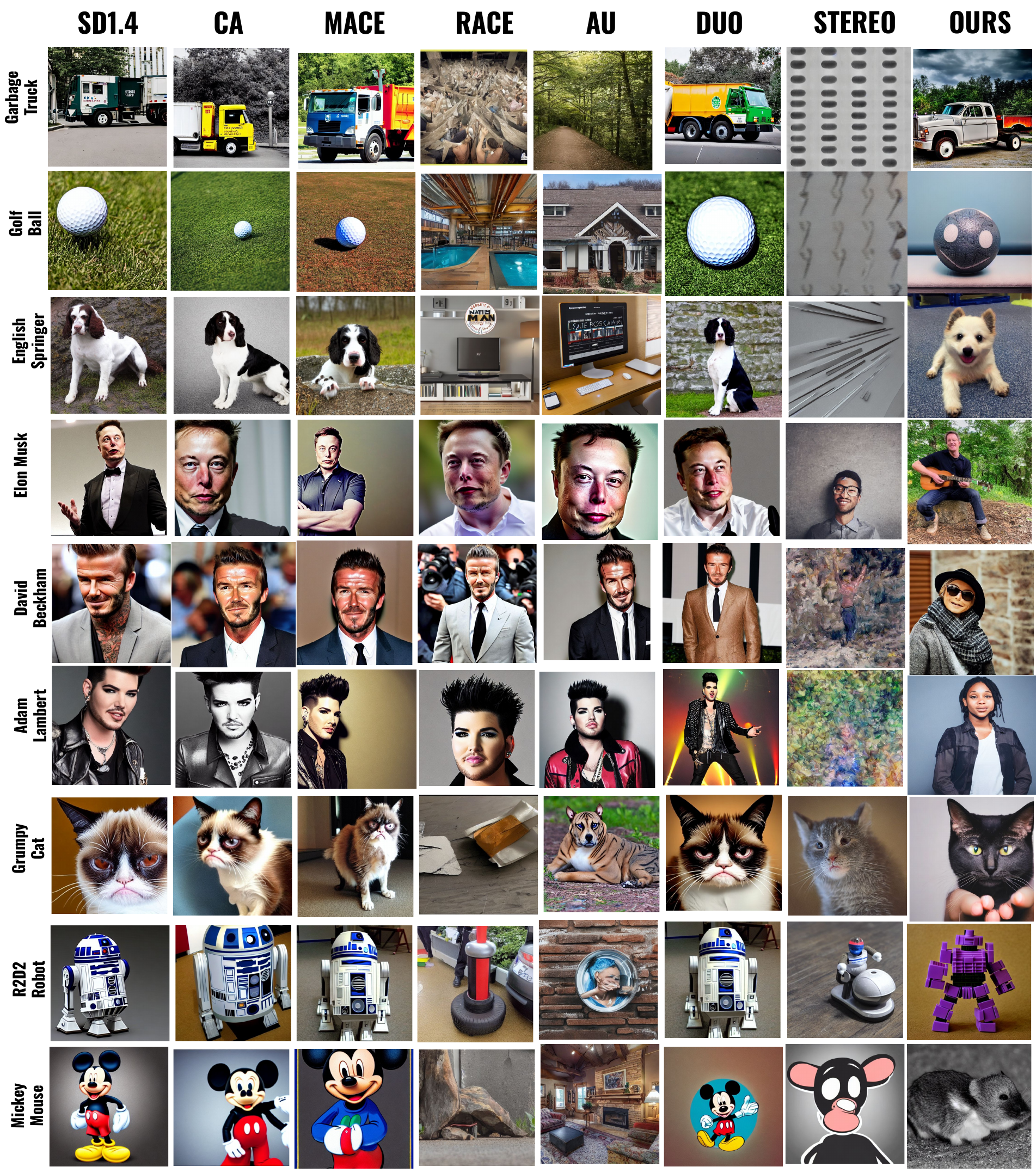}
   \caption{Result of UnlearnDiff (UD) Attack under diffirent methods.} 
   \label{fig:UD}
\end{figure*}

We present the output of the CCE attack and the normal target's textual representation for the remaining concepts not included in the main paper in Figures \ref{fig:Remain Attack Output} and \ref{fig:Remain Textual Output}, respectively, as well as the results of different methods under the UnlearnDiff (UD) attack in Figure \ref{fig:UD}. For the CCE attack, all methods—CA, MACE, RACE, and AU—perform similarly to the concepts presented in the main paper, and all are successfully attacked. For STEREO, it performs well for \textit{Golf ball} and \textit{David Beckham}, but its outputs for the remaining concepts are meaningless. Our method, on the other hand, erases the main attributes of the target concept while preserving its general meaning. This demonstrates that our method achieves a balance between \textit{completeness} and \textit{effectiveness}.

